\documentclass[letterpaper, 10 pt, journal, twoside]{IEEEtran}

\usepackage{amsmath,amsfonts,amssymb}
\usepackage{array}
\usepackage[caption=false,font=normalsize,labelfont=sf,textfont=sf]{subfig}
\usepackage{textcomp}
\usepackage{stfloats}
\usepackage{url}
\usepackage{verbatim}
\usepackage{graphicx}
\usepackage{balance}
\def\BibTeX{{\rm B\kern-.05em{\sc i\kern-.025em b}\kern-.08em
    T\kern-.1667em\lower.7ex\hbox{E}\kern-.125emX}}
\usepackage{balance}
\usepackage{xcolor}
\usepackage{color,soul}
    \definecolor{codegreen}{rgb}{0,0.6,0}
    \definecolor{codegray}{rgb}{0.5,0.5,0.5}
    \definecolor{codepurple}{rgb}{0.58,0,0.82}
    \definecolor{backcolour}{rgb}{0.95,0.95,0.92}

\usepackage[isbn=false,
    backend=biber,
    style=ieee,
    bibstyle=ieee,  
    citestyle=numeric-comp,     
    sortcites=true,   
    giveninits=true,            
    backref=false]{biblatex}
\usepackage{hyperref}
\usepackage{algorithm}
\usepackage{algpseudocode}
\usepackage[dvipsnames]{xcolor}

\usepackage{booktabs} 
\usepackage{bbm} 
\usepackage{float}
\usepackage{stfloats}

\newcommand{\norm}[1]{\left\lVert#1\right\rVert}

\DeclareMathOperator{\xvec}{\mathbf{x}}
\DeclareMathOperator{\uvec}{\mathbf{u}}
\DeclareMathOperator{\qvec}{\mathbf{q}}
\DeclareMathOperator{\qdvec}{\mathbf{\dot{q}}}
\DeclareMathOperator{\qddvec}{\mathbf{\ddot{q}}}
\DeclareMathOperator{\Fvec}{\mathbf{F}}
\DeclareMathOperator{\zvec}{\mathbf{z}}
\DeclareMathOperator{\vvec}{\mathbf{v}}

\begin{document}


\title{Residual MPC: Blending Reinforcement Learning \\
       with GPU-Parallelized Model Predictive Control}
\author{Se Hwan Jeon$^1$, Ho Jae Lee$^1$, Seungwoo Hong$^2$, Sangbae Kim$^1$
    \thanks{$^1$Department of Mechanical Engineering, Massachusetts Institute of Technology, Cambridge MA, USA.}
    \thanks{$^1$Department of Mechanical Engineering, School of Smart Mobility, Korea University, Seoul, South Korea.}
}
\maketitle

\newcommand{\customCommand}{\bf{example custom command}}
\newcommand{\mithumanoid}{the MIT Humanoid~\cite{saloutos2023design}}
\newcommand{\done}{{\color{codegreen}{\checkmark}}}

\renewcommand{\figureautorefname}{Fig.}

\newcommand\blfootnote[1]{%
  \begingroup
  \renewcommand\thefootnote{}\footnote{#1}%
  \addtocounter{footnote}{-1}%
  \endgroup
}

\newcommand{\sehwan}[1]{\textcolor{blue}{Se Hwan: #1}}
\newcommand{\hojae}[1]{\textcolor{orange}{HoJae: #1}}

\begin{abstract}
Model Predictive Control (MPC) provides interpretable, tunable locomotion controllers grounded in physical models, but its robustness depends on frequent replanning and is limited by model mismatch and real-time computational constraints.
Reinforcement Learning (RL), by contrast, can produce highly robust behaviors through stochastic training but often lacks interpretability, suffers from out-of-distribution failures, and requires intensive reward engineering.
This work presents a GPU-parallelized residual architecture that tightly integrates MPC and RL by blending their outputs at the torque-control level.
We develop a kinodynamic whole-body MPC formulation evaluated across thousands of agents in parallel at 100 Hz for RL training.
The residual policy learns to make targeted corrections to the MPC outputs, combining the interpretability and constraint handling of model-based control with the adaptability of RL.
The model-based control prior acts as a strong bias, initializing and guiding the policy towards desirable behavior with a simple set of rewards.
Compared to standalone MPC or end-to-end RL, our approach achieves higher sample efficiency, converges to greater asymptotic rewards, expands the range of trackable velocity commands, and enables zero-shot adaptation to unseen gaits and uneven terrain.
\blfootnote{Video: https://www.youtube.com/watch?v=L2NrPD4yiMs}
\end{abstract}

\begin{figure*}[tb]
  \centering
  \includegraphics[width=\textwidth]{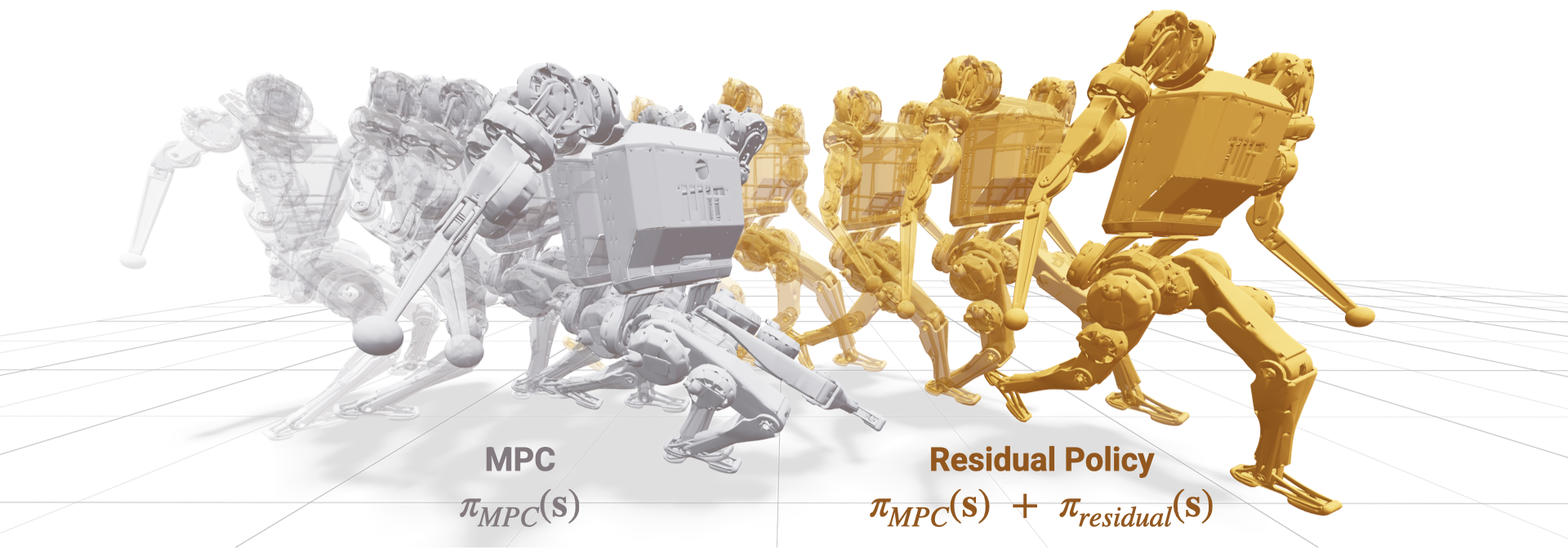}
  \caption{
      Simulated locomotion for MPC (left, grey) and the residual policy (right, orange) at velocity commands of 2.25 m/s and -1.5 rad/s.
      The MPC controller fails and terminates from self-collision, while the residual policy is able to stabilize the robot.
      We use Viser for 3D visualization in this work~\cite{yi2025viser}.
  }
  \label{fig:mpc_vs_residual}
\end{figure*}

\section{Introduction}

\IEEEPARstart{M}{odel} predictive control (MPC) has been widely used for robotic locomotion control due to its grounding in physical models of the robot’s dynamics and constraints~\cite{hong2020real,ding2021representation,di2018dynamic,khazoom2024tailoring,dai2014whole,garcia2021mpc}.
By explicitly predicting the evolution of the system over a finite horizon, MPC can generate control inputs that optimize performance while respecting physical limits such as joint torques, contact forces, and stability margins.
These models are valid throughout the system's state space and, because the underlying optimization problem is explicit, they provide interpretable predictions and tunable parameters for transparent controller design.
The designer can readily inspect internal quantities such as predicted trajectories, constraint satisfaction, or cost terms, which makes MPC attractive not only for deployment but also for debugging and analysis.
However, the performance of MPC derives primarily from frequent replanning and online optimization, which makes it vulnerable to inaccuracies in the assumed model.
Even small discrepancies between the modeled and actual system, such as uncertainties in terrain shape, variable ground friction, actuator nonlinearities, or unmodeled compliance, can accumulate between replanning steps and destabilize the controller, particularly with computation or actuation delays.
This sensitivity to modeling errors and environmental disturbances remains a key challenge for pure model-based locomotion control.
Moreover, because MPC relies on solving a structured optimization problem at high bandwidth, the need for real-time, gradient-based computation also restricts the expressiveness of the models and constraints that can be incorporated in practice~\cite{shen2022convex,khazoom2024tailoring,di2018dynamic}.

In contrast, reinforcement learning (RL) provides a complementary paradigm that bypasses the need for explicit modeling by directly optimizing control policies from simulated data.
By training stochastically under diverse simulated conditions, RL can expose policies to a broad spectrum of uncertainties, perturbations, and noise, producing behaviors that often exhibit remarkable robustness to parameter uncertainty and the ability to generalize to unseen terrains~\cite{miki2022locomotion,gu2024humanoid,radosavovic2024real,batke2022optimizing,xie2020learning,xie2018feedback}.
Policies learned through RL can, in principle, adapt to phenomena that are difficult to model, such as unexpected contact events or nonlinear actuator dynamics~\cite{siekmann2021stairs,hwangbo2019actuator}.
Nonetheless, RL methods are hampered by the black-box nature of neural networks, which limits interpretability and complicates debugging or analysis: it is often unclear why a given action was chosen, or how a policy will behave outside the training distribution.
Furthermore, the design of rewards, penalties, or curricula to steer learning toward a desired behavior can often be unintuitive and time-consuming.
Small changes in reward weighting or environment randomization can lead to drastically different policies, sometimes exploiting numerical simulation artifacts rather than learning physically plausible locomotion behavior~\cite{randlov1998bicycle,openai2016faulty}.

Blending MPC and RL seeks to combine the interpretability, tunability, and predictive guarantees of MPC with the robustness and adaptability of RL.
The intuition is to let MPC provide a structured, physically consistent baseline while allowing a learned policy to make targeted corrections that account for unmodeled effects or constraints difficult to include in the formulation.
This integration offers the potential to reduce the dependence on reward engineering, manual curriculum design, and hyperparameter tuning that is characteristic of end-to-end RL.
Various architectures have been proposed to achieve such a synthesis, including hierarchical structures where a high-level policy outputs MPC parameters (e.g., ~\cite{romero2024actor}), learning to track predicted states and commands from MPC predictions (e.g., ~\cite{jenelten2024dtc}), or embedding learnt dynamics or constraints into an MPC formulation (e.g., ~\cite{lutter2021learning}).
A thorough survey of synthesizing MPC and RL is presented in~\cite{reiter2025synthesis}.

Another promising direction is to adopt a parallel or \textit{residual} policy architecture, where the outputs of a fixed control prior are modified with corrective actions from a trained policy~\cite{silver2018residual,luo2024residual,johannink2019residual}.
Whereas a hierarchical architecture depends on reasonable outputs from the higher-level controller, a parallel architecture can have its components interact with each other adaptively.
From the perspective of model-based control, the residual network would serve as a corrective layer for model-based outputs - given model simplifications or uncertainties in assumed parameters such as the ground height, environment friction, or inertial parameters, a learned layer could accordingly adjust the model-based torques.
MPC formulations are typically sensitive to parameter inputs such as noise or contact timing, and mismatches can quickly cause solutions to diverge.
RL policies, on the other hand, can be trained to be robust to these uncertainties.
A learned, adaptive layer could retain the interpretability and tunability of the core model-based controller, while adapting to erroneous model assumptions or changing environmental conditions.

From the perspective of reinforcement learning, the fixed control prior in the residual architecture can be considered as a ``warm start" for the policy.
In the RL setting, we can freely design any number of sparse, nondifferentiable, and/or complex rewards as desired, and with the residual architecture, initialize the search for a satisfactory optimal policy with a model-based controller.
By starting ``closer" to an optimal policy, training is more sample efficient as the control prior guides the agents towards desirable regions of state space~\cite{silver2018residual}.

However, embedding MPC directly into a learning pipeline introduces a severe computational bottleneck.
Training residual policy would require solving high-dimensional optimizations for each of the thousands of simulated agents necessary for RL.
While prior work has explored training residual policies with other underlying controllers, including linear-quadratic regulators (LQRs), pretrained policies, and instantaneous whole-body controllers~\cite{cramer2024contextualized, youm2023imitatingMPC, cheng2025rambo, luo2024residual, silver2018residual, johannink2019residual}, these are much smaller problems, operating only on a single timestep.
An MPC formulation for the same system optimizes over the entire prediction horizon while enforcing dynamics constraints, dramatically increasing the problem dimensionality and complexity.
Due to the prohibitive computation required, relatively few works have explored training a policy alongside an MPC controller.

\textcite{jenelten2024dtc} accomplished this by leveraging large scale CPU clusters to evaluate many instances of the MPC in parallel, but this was only possible by solving the MPC at 2-3 Hz during training.
With the additional computational overhead of transferring data between the CPU and GPU, training required weeks for full convergence.
The proposed controller was also hierarchical rather than residual, where the policy was trained to track the predicted MPC states and foot commands rather than modify them in real-time.
In contrast, our work investigates the effect of including concurrent, real-time MPC within an RL training loop at the torque-level.

Instead of performing MPC solves and RL iterations on separate devices, recent parallelization tools such as \texttt{CusADi} and \texttt{cuDSS} can be leveraged to efficiently solve many MPC instances concurrently on the GPU so that all memory is shared, eliminating any overhead from data transfer~\cite{jeon2024cusadi, cudss}.
This enables high-frequency MPC solutions to be integrated into the RL training loop without the prohibitive latency and memory transfer costs observed in prior work.

\begin{figure*}[tbp]
  \centering
  \includegraphics[width=0.8\textwidth]{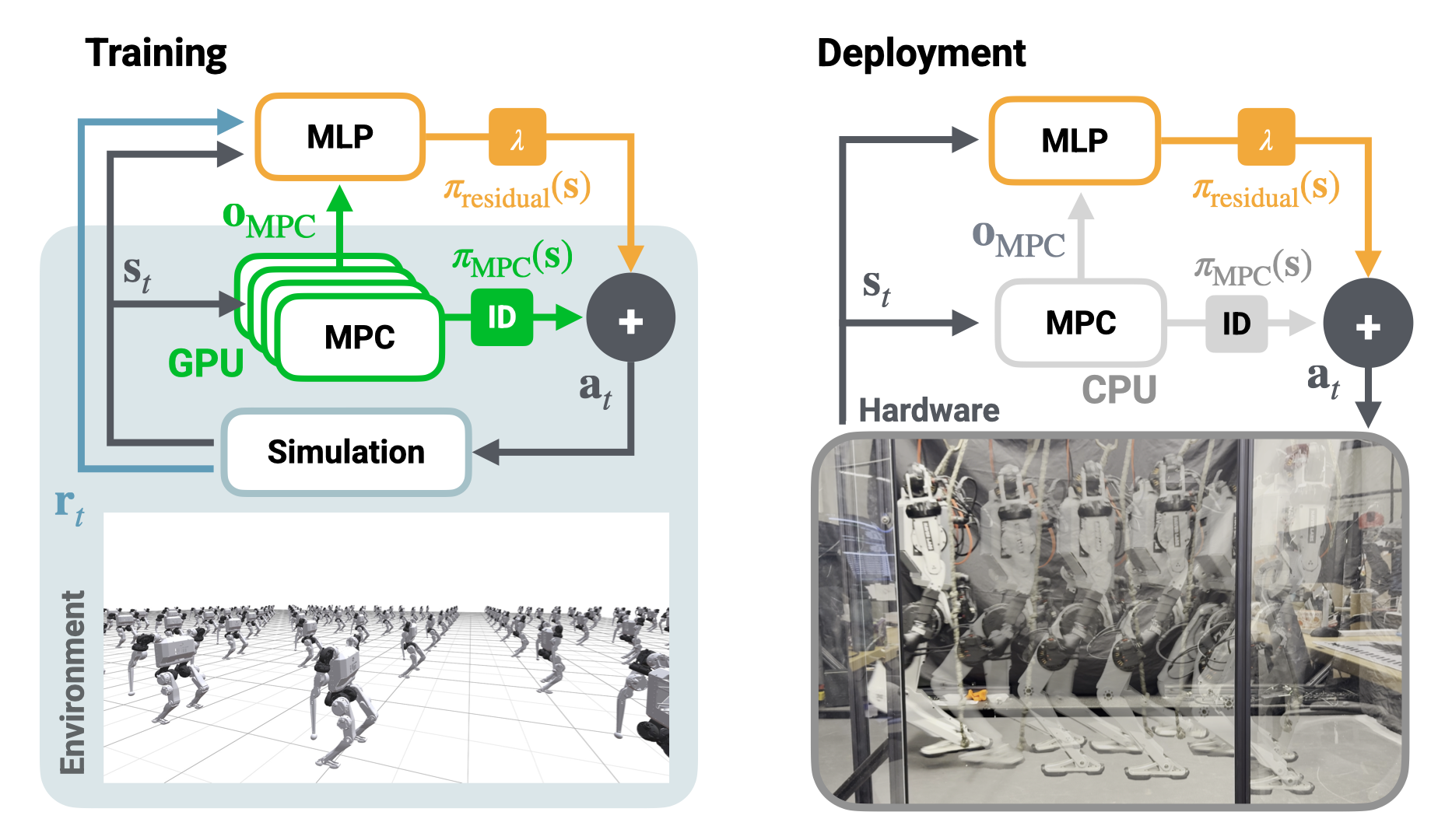}
  \caption{
    Diagram of the proposed residual policy architecture during training and deployment.
    Both the multilayer perceptron (MLP) policy and MPC are evaluated at the control frequency of 100 Hz, and are blended downstream with a scaling factor of $\lambda$ on the policy output.
    The MPC solution is converted into desired torques, joint positions, and joint velocities with the inverse dynamics (ID) of the system.
    We efficiently evaluate the MPC controller across thousands of instances on the GPU for concurrent training with the residual network.
  }
  \label{fig:diagram}
\end{figure*}

For our work, we study the effects of pairing a residual policy architecture with a whole-body MPC controller, where the outputs of a trained network and predictive controller are directly blended at the torque-control level, as shown in \autoref{fig:diagram}.
We introduce a kinodynamic whole-body MPC formulation that is parallelized for high-frequency GPU evaluation and is solved concurrently with the residual network at 100 Hz in IsaacGym~\cite{makoviychuk2021isaac}.
To the authors' knowledge, this represents the first work for legged systems embedding high-frequency MPC at the torque level within an RL training loop. 
With this setup, the residual policy outperforms both standalone MPC and baseline end-to-end RL controllers.

This residual training exhibits substantially higher sample efficiency and converges to a greater asymptotic reward than end-to-end RL.
By providing a strong MPC control prior, the residual architecture biases the policy towards realistic and desirable locomotion behaviors, whereas end-to-end RL tends to exploit simulation idiosyncrasies and produce aphysical motions.
This substantially reduces the need for the arduous reward engineering and tuning that typically burden standard RL training.

From a performance standpoint, the residual policy markedly extends the range of trackable velocity commands beyond the capabilities of the baseline MPC, as shown in \autoref{fig:mpc_vs_residual}.
It also learns to modify the MPC outputs to satisfy complex constraints such as self-collision, which are challenging to encode in model-based formulations without system-specific heuristics.
For instance, prior work has relied on constraining distances between geometric primitives throughout the MPC horizon, or using lower-level whole-body optimizations to prevent collision~\cite{khazoom2024tailoring,khazoom2022humanoid}.
With RL, it is much easier to define sparse, non-differentiable rewards, and learn to avoid self-collision through data-driven experience.
Additionally, we find that the residual policy exhibits zero-shot adaptability to unseen changes in the requested gait and uneven terrain.
While MPC fails under contact plans with double stance or flight phases, the residual policy is able to track these trajectories reliably.
Similarly, over uneven terrain, the MPC baseline quickly fails due to incorrect assumptions about contact timings, whereas the residual policy successfully navigates the variations of the ground.

Finally, we examine how the learned corrections from the residual network interact with the MPC prior, and interpret the residual actions with respect to the concurrent MPC torques.
Interestingly, we find that the residual torques are generally antagonistic to the MPC torques, particularly shortly before a planned touchdown contact.
Examining these corrective torques suggests that the residual policy heavily engages the ankle near touchdown and takeoff.
We hypothesize that this behavior may explain the robustness present in learned locomotion policies with respect to making and breaking contact, and could serve as a heuristic for designing controllers in the future.

We summarize our contributions as follows:
\begin{itemize}
    \item Embedding a GPU-parallelized, high-frequency MPC controller into RL environments for in-the-loop training.
    \item Analyzing how the MPC control prior guides and interacts with the residual network during locomotion.
    \item Demonstrating zero-shot adaptation of the residual policy to changes in the MPC parameters.
    \item Validating the residual policy on hardware for \mithumanoid.
\end{itemize}

\section{Background}
\subsection{Model Predictive Control}
\label{sec:mpc_background}
MPC is a model-based control technique that optimizes a cost function over a finite horizon, subject to system dynamics and constraints.
By repeatedly solving and updating this optimization problem at high frequency, the controller can adapt its actions to changing conditions and disturbances.

We denote the discrete, finite-time optimal control problem generally as
\begin{eqnarray}
\label{eq:OCP}
    \min_{\xvec[\cdot], \uvec[\cdot]}   && l_T(\xvec_T, \uvec_T) + \sum_{i=0}^{T-1} l_i(\xvec_i, \uvec_i) \\
    \text{s.t.}             && \xvec_0 = \bar{\xvec}_0 \nonumber \\
                            && \xvec_{i+1} = \mathbf{f}(\xvec_i, \uvec_i), \quad i=0,\dots, T-1 \nonumber \\
                            && \mathbf{g}_i(\xvec_i, \uvec_i) \leq 0, \quad i=0, \dots, T \nonumber
\end{eqnarray}
where $\xvec[\cdot] \in \mathbb{R}^{n_x \times T}$ and $\uvec[\cdot] \in \mathbb{R}^{n_u \times T}$ are the state and control trajectories with initial condition $\bar{\xvec}_0$ and $\mathbf{f}: \mathbb{R}^{n_x} \times \mathbb{R}^{n_u} \rightarrow \mathbb{R}^{n_x}$ is the nonlinear dynamics of the system.
The system is constrained by $\mathbf{g}:\mathbb{R}^{n_x} \times \mathbb{R}^{n_u} \rightarrow \mathbb{R}^{m}$, and optimized for the costs $l:\mathbb{R}^{n_x} \times \mathbb{R}^{n_u} \rightarrow \mathbb{R}$ summed across the timesteps $i$.

This nonlinear program (NLP) can be solved effectively with sequential quadratic programming (SQP) approaches~\cite{Boyd2004_ConvexOptimization}.
By iteratively solving subproblems of the original NLP with a quadratic model of the cost and subject to linearized constraints, initial guesses to \eqref{eq:OCP} can quickly converge to an optimal solution.

Under computational limits, the so-called \textit{real-time iteration} scheme is an effective technique to quickly approximate an optimal solution~\cite{Diehl2005_realtimeIteration}.
The scheme involves strictly limiting the number of subproblems solved, directly trading optimality for responsiveness.
For real-time systems with fast dynamics, solving even a single subproblem can be sufficient for stable closed-loop performance while greatly reducing computational effort~\cite{Diehl2005_realtimeIteration, jeon2024cusadi}.

\subsection{Residual Policy Learning}
Standard deep reinforcement learning methods are initialized with arbitrary policies $\pi$ that are trained to maximize the expected return of a task, given as
\begin{equation}
    J(\pi) = \mathbb{E}_{\tau \sim \pi}\left[\sum_{t=0}^{\infty} \gamma^t r(s_t, a_t)\right],
    \label{eq:rl_return}
\end{equation}
where $\tau = (s_0, a_0, s_1, a_1, \ldots)$ is a state-action trajectory sampled from $\pi$, $\gamma$ is the discount factor, and $s_t$, $a_t$, and $r(s_t, a_t)$ are the state, action, and reward received at time $t$ respectively.

A residual network (ResNet) is a neural network with a "shortcut" or "skip" connection, that allows the input to be added directly to the output of a series of layers~\cite{he2016deep}.
When the output of a desired function is close to the input, a residual network structure can significantly improve the convergence speed and accuracy of training in a supervised context.
As one example, discrete dynamical systems are well-suited for a residual structure, as the evolution of the state is typically a small perturbation from the previous state, $x_{k+1} = x_k + f(x_k)$.
This has been explored deeply in the context of \textit{physics-informed neural networks} (PINNs), where the residual architecture can be used to learn complex dynamics such as fluid flow or chemical reactions~\cite{raissi2019physics, lu2021deepxde}.

This idea has been extended to learning residuals of \textit{policies}, rather than individual inputs - given a nominal controller $\pi_0$, a residual policy $\pi_{\mathrm{residual}}$ can be trained to maximize \eqref{eq:rl_return}, where trajectories are drawn from samples of $\pi = \pi_0 + \pi_{\mathrm{residual}}$.
This approach has been shown to substantially improve sample efficiency, convergence speed, and asymptotic performance of policies in a variety of tasks~\cite{johannink2019residual,silver2018residual,cramer2024contextualized}.
Here, we focus on adapting and improving locomotion with this architecture, aiming to augment the performance of MPC with a learned residual layer, as shown in \autoref{fig:diagram}.

\section{MPC Formulation and Parallelization}
For designing MPC controllers in general, care needs to be taken to balance model fidelity with computational constraints.
A formulation equipped with accurate dynamics and detailed costs/constraints may be severly limited by the time required for a solution, limiting its replanning rate and overall responsiveness.
Conversely, a simplified model may be solved quickly, but could make unrealistic predictions that lead to poor or limited closed-loop performance.
This consideration is especially important for GPU parallelization, where a learning iteration requires solving many MPC problems for each trajectory rollout in a batch - complex formulations can greatly increase training times, requiring days or even weeks for policies to converge~\cite{romero2024actor, jenelten2024dtc}.

The equations of motion for a floating-base robot with generalized coordinates $\qvec \in \mathbb{R}^n$ and generalized velocities $\qdvec \in \mathbb{R}^n$ can be written as
\begin{equation}
    \mathbf{M}(\qvec)\qddvec + \mathbf{h}(\qvec, \qdvec) = \mathbf{J}(\qvec)^\intercal \Fvec + \mathbf{S}^\intercal \boldsymbol{\tau},
\end{equation}
where $\mathbf{M}(\qvec) \in \mathbb{R}^{n \times n}$ is the mass matrix, $\mathbf{h}(\qvec, \qdvec) \in \mathbb{R}^{n}$ is the vector of Coriolis, centrifugal, and gravitational forces, $\mathbf{J}(\qvec) \in \mathbb{R}^{3n_c \times n}$ is the contact Jacobian for $n_c$ contact points, $\Fvec \in \mathbb{R}^{3n_c}$ is the vector of contact forces, $\mathbf{S} \in \mathbb{R}^{n_j \times n}$ is a selection matrix mapping actuated joint torques to generalized forces, and $\boldsymbol{\tau} \in \mathbb{R}^{n_j}$ is the vector of actuated joint torques.
These system dynamics are the primary bottleneck for MPC problems, and many formulations focus on reducing its complexity to be amenable for real-time control~\cite{garcia2021mpc, khazoom2024tailoring, di2018dynamic,ding2021representation, hong2020real}

Our system is \mithumanoid, weighing roughly 25 kg in total with ten actuated joints for the legs and eight for the arms.
For this work, we only use proprioceptive measurements from the internal IMU and joint encoders, and do not rely on contact sensors or cameras for locomotion.
We define $n_c = 4$ contact points for our system, corresponding to the right toe, right heel, left toe, and left heel for the contact Jacobians.

\subsection{Formulation}
\label{sec:mpc_formulation}
\begin{figure}[tbp] 
    \centering
    \includegraphics[width=\columnwidth]{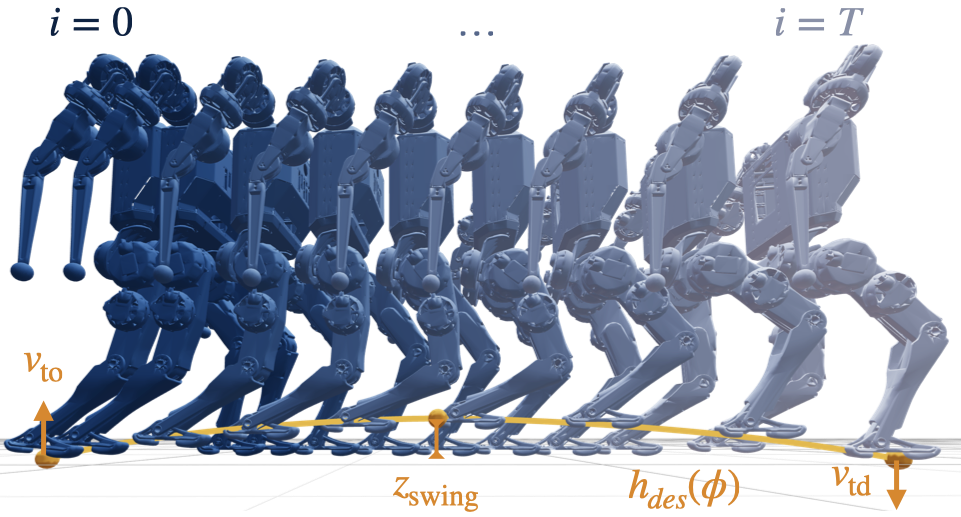}
    \caption{
        Visualized predictions for the MPC formulation (distances are adjusted for visual clarity).
        A Bezier curve is defined by the peak swing height, takeoff velocity, and touchdown velocity.
        This curve is parametrized by the contact phase to output desired foot heights.
    }
    \label{fig:mpc_breakdown}
\end{figure}

For our work, we formulate a ``floating-base kinodynamic" optimization as a balance between overly simplified single rigid-body models and a computationally prohibitive whole-body MPC formulation.
Only the dynamics of the floating base are considered by multiplying the equations of motion with a selection matrix $\mathbf{S}_{base} = [\mathbf{I}_{n_b \times n_b} \; \mathbf{0}_{n_b \times n_j}]$, where $n_b$ is the number of base coordinates and $n_j$ is the number of joints.
This reduces the equations of motion to
\begin{equation}
    \mathbf{M}_{b}(\qvec)\qddvec + \mathbf{h}_{b}(\qvec, \qdvec) = \mathbf{J}_{b}(\qvec)^\intercal \Fvec,
    \label{eq:base_dynamics}
\end{equation}
which serves as a path constraint of dimension $n_b$ on $\qvec$, $\qdvec$, $\qddvec$,and $\Fvec$\footnote{Projecting \eqref{eq:base_dynamics} to the center of mass frame is exactly the \textit{centroidal dynamics} formulation used in prior work \cite{Orin2013_centroidal, dai2014whole, wensing2016improved, garcia2021mpc}.
We found that there was little difference in closed-loop performance between these formulations, while the base dynamics formulation was slightly faster to compute.}.

With this dynamics constraint, we choose to optimize over the generalized coordinates $\qvec$, velocities $\qdvec$, and ground reaction forces $\Fvec$, over a horizon length of $T = 12$.
Writing these decision variables as $\zvec := [\qvec^\intercal \; \qdvec^\intercal \; \Fvec^\intercal]^{\intercal}$, our MPC formulation is
\begin{eqnarray}
    \label{eq:mpc_formulation}
    \min_{\qvec[\cdot], \qdvec[\cdot], \Fvec[\cdot]} && \sum_{i=0}^T(\zvec_i - \zvec_{i, des})^\intercal Q_{\zvec} (\zvec_i - \zvec_{i, des})\Delta t_i \\
    \text{s.t.}  && \; \forall \; i=0, \dots, T-1 \nonumber \\
    \text{(Initial state)}  && \qvec_0 = \bar{\qvec}_0, \nonumber \\
                            && \qdvec_0 = \bar{\dot{\mathbf{q}}}_0, \nonumber \\
    \text{(Integration)}    && \qdvec_{i+1} = \qdvec_i + \Delta t_i \qddvec_{i}, \nonumber \\
                            && \qvec_{i+1} = \qvec_i + \Delta t_i \qdvec_{i+1}, \nonumber \\ 
    \text{(Dynamics)}       && \mathbf{M}_{b}(\qvec_i)\qddvec_i + \mathbf{h}_{b}(\qvec_i, \qdvec_i) = \mathbf{J}_{b}(\qvec_i)^\intercal \Fvec_i, \nonumber \\ 
                            && \qddvec_i = (\qdvec_{i+1} - \qdvec_{i})/{\Delta t_i}, \nonumber \\
    \text{(Contact)}        && \norm{\Fvec_{x,y}} \leq \mu \Fvec_z \cdot \boldsymbol{\phi}_i, \nonumber \\
                            && \mathbf{J}(\qvec_i)\qdvec_i \cdot \boldsymbol{\phi}_i = \mathbf{0}, \nonumber \\
    \text{(Swing)}          && \mathbf{r}(\qvec_i)_z = h_{i, des}, \nonumber \\
    \text{(Joint limits)}   && \qvec_{i, joint} \in \mathcal{Q}, \nonumber  \\
                            && \qdvec_{i, joint} \in \dot{\mathcal{Q}}, \nonumber
\end{eqnarray}
where $Q_{\zvec} \succeq 0$ is a diagonal weight matrix, $\zvec_{i, des}$ is a desired state, $\Delta t_i$ is the timestep duration, $\bar{\qvec}_0$ and $\bar{\dot{\mathbf{q}}}_0$ are the current state of the robot, $\mu$ is the static coefficient of friction, $\boldsymbol{\phi}_i$ is the contact phase, $\mathbf{r}(\qvec_i)_z$ is the foot height, $h_{i, des}$ is the desired foot height, and $\mathcal{Q}$ and $\dot{\mathcal{Q}}$ are the feasible set of joint position and velocity limits.

For the desired state trajectory $\zvec_{des}[\cdot]$, we set the desired vertical ground reaction forces equal to the system's weight divided by the number of contacts.
For $\qvec_{i,des}$ and $\qdvec_{i,des}$, we use a simple extrapolation heuristic based on the MPC command $\mathbf{c} \in \mathbb{R}^4$ .
The command consists of the desired height $c_h$, linear velocities $c_{v_x}$ and $c_{v_y}$, and yaw angular velocity $c_{\omega_z}$, and we compute the desired trajectories as follows:
\begin{align}
    q^h_{i,des} &= c_h, \label{eq:des_traj_1} \\
    q^{\theta_z}_{i,des} &= q^{\theta_z}_{i-1,des} + \Delta t_i c_{\omega_z},  \\
    \dot{q}^{v_x}_{i,des} &= c_{v_x}, \\
    \dot{q}^{v_y}_{i,des} &= c_{v_y}, \\
    \dot{q}^{\omega_z}_{i,des} &= c_{\omega_z},\label{eq:des_traj_5}
\end{align}
for $i=0, \dots, T-1$.
The desired joint positions are set to a nominal position for each timestep, and all other desired quantities are set to zero.
To avoid sensitivity to absolute position drift, the weights on the $x$ and $y$ states are set to zero as well.

We design a fixed contact schedule with phase variable $\boldsymbol{\phi} \in \mathbb{R}^{n_c}$ for each contact point.
For our work, we set the period to be 0.8s and enforce contact switches from stance to swing when $\boldsymbol{\phi} = 0.5$, but these can easily be adjusted to command double stance or flight phases, as described in~\cite{bledt2018contact}.
This phase variable is integrated through the predicted horizon of the MPC, and each node is assigned a corresponding contact state based on whether the phase is larger or smaller than the switching value.
Our controller fixes a phase offset of 0.5 between the left and right contact points, commanding the humanoid to step flat footed without heel-toe transitions.

The foot height is constrained to a curve $h_{i, des}$ based on the duration of swing, as shown in \autoref{fig:mpc_breakdown}.
This curve is formulated as a fifth-order Bezier curve with the following conditions:
\begin{align}
    B(t_{sw}=0.0) &= 0, \label{eq:bezier_1}\\
    B(t_{sw}=1.0) &= 0, \\
    B(t_{sw}=0.5) &= z_{\mathrm{swing}}, \\
    \dot{B}(t_{sw} = 0.0) &= v_{\mathrm{to}} \\
    \dot{B}(t_{sw} = 1.0) &= v_{\mathrm{td}} \label{eq:bezier_5}
\end{align}
where $t_s$ is the time during swing, scaled to be between 0 and 1, and $v_{\mathrm{to}}$, $v_{\mathrm{td}}$, and $z_{\mathrm{swing}}$ are the desired takeoff velocity, touchdown velocity, and swing height respectively.

The formulation in \eqref{eq:mpc_formulation} sacrifices the ability to reason about individual joint dynamics and torques, limiting the controller's predictions for dynamic motions close to the actuation limits of the robot.
Additionally, we do not have any bounds on the floating base states $\qvec_{i, base}$ and $\qdvec_{i, base}$, or self-collision constraints as in \cite{khazoom2024tailoring}, as these can further increase computational complexity.
With the residual policy architecture, we show that these limitations can be overcome by the learned policy, and the controller combined is capable of avoiding self-collisions and termination.

\subsection{Solver Strategy}
As discussed in \autoref{sec:mpc_background}, we use the \textit{real-time iteration} scheme to solve the nonlinear optimization with a single SQP iteration per MPC solve.
This reduces \eqref{eq:mpc_formulation} to a single quadratic program (QP) to solve, which can be approached with various active-set, interior-point, or alternating direction implementations~\cite{pandala2019qpswift,ferreau2014qpoases,stellato2020osqp,schwan2023piqp}.

We adopt the techniques in the Operator-Splitting Quadratic Program (OSQP) library~\cite{stellato2020osqp} to solve \eqref{eq:mpc_formulation}.
Compared to other QP solution methods, the alternating direction method of multipliers (ADMM) used in OSQP requires the fewest symbolic computations per iteration.
Interior point and active-set methods modify the KKT matrix at each iteration (from updating constraint Jacobians and/or the barrier parameter), requiring a full refactorization of the system at each step.
The KKT matrix is seldom modified during solves with ADMM, and subsequent iterations consist only of simple projection and matrix-vector operations.

The most expensive step is factorizing and solving the KKT system of the QP subproblem, given as 
\begin{equation}
    \label{eq:kkt_system}
    \underbrace{
        \begin{bmatrix} \mathbf{P} + \sigma \mathbf{I} & \mathbf{A}^\intercal \\ \mathbf{A} & -\rho^{-1}\mathbf{I} \end{bmatrix}}_{\mathbf{A}_K
        } \vvec_K 
    =
    \underbrace{
        \begin{bmatrix} \sigma \mathbf{x}_{\mathrm{QP}} - \mathbf{q} \\ \mathbf{z}_{\mathrm{QP}} - \rho^{-1}\mathbf{y}_{\mathrm{QP}} \end{bmatrix}
        }_{\mathbf{b}_K},
\end{equation}
where $\mathbf{P}$, $\mathbf{q}$, and $\mathbf{A}$ are the cost Hessian\footnote{Note that the cost of \eqref{eq:mpc_formulation} is in linear least-squares form, and the Gauss-Newton Hessian approximation is equivalent to the full Hessian.}, the cost Jacobian, and the constraint Jacobian of the MPC formulation \eqref{eq:mpc_formulation} respectively, and $\mathbf{I}$ is the identity matrix.
$x_\mathrm{QP}$ and $y_\mathrm{QP}$ are the current primal and dual variables of the QP subproblem, $\sigma$ is the ADMM penalty parameter, and $\rho$ is the dual variable scaling factor.

To solve the system, the first step involves factorizing the KKT matrix $\mathbf{A}_K$
\begin{equation}
    \mathbf{L}_K \mathbf{D}_K \mathbf{L}_K^\intercal = \mathbf{A}_K, \label{eq:ldl}
\end{equation}
into a lower-triangular matrix $\mathbf{L}_K$ and a diagonal matrix $\mathbf{D}_K$.
For a single QP solve, this LDL factorization only needs to be performed once.
The subsequent iterations of the QP are performed by updating $\mathbf{b}_K$, and performing back-substitution to solve for $\vvec_K$:
\begin{align}
    \mathbf{L}_K \mathbf{D}_K \mathbf{L}_K^\intercal \vvec_K = \mathbf{b}_K, \label{eq:backsolve_1}\\
    \vvec_K = \mathbf{L}_K^{-\intercal} \mathbf{D}_K^{-1} \mathbf{L}_K^{-1} \mathbf{b}_K. \label{eq:backsolve_2}
\end{align}
Evaluating \eqref{eq:backsolve_2} is computationally inexpensive, requiring only diagonal scaling and forward/backward substitution.

After the system is solved for $\mathbf{v}_K$, $\mathbf{b}_K$ and the primal/dual variables are updated with simple projection and linear algebra operations.
These operations and the linear solve \eqref{eq:backsolve_2} are then repeated for a fixed number of QP iterations $N_{\mathrm{QP}}$, or until convergence.
For details on these computations, readers are referred to Section 3.2 in~\cite{stellato2020osqp}.

\subsection{MPC Computation}
Overall, the MPC controller is evaluated as follows.
First, an initial guess $\zvec[\cdot]$ is computed based on the current state of the robot and the desired command with $f_{\mathrm{init}}$.
This consists of some nominal pose repeated for every timestep of the trajectory, with all other quantities set to zero.
The parameters of the MPC problem $\boldsymbol{\theta}_{\mathrm{MPC}}$ such as the desired states, contact schedule, and commands are then updated with $f_{\mathrm{param}}$, consisting of \eqref{eq:des_traj_1}-\eqref{eq:des_traj_5} and \eqref{eq:bezier_1}-\eqref{eq:bezier_5}.

Next, the KKT system is constructed with $f_{\mathrm{KKT}}$, which computes the necessary Hessians and Jacobians of \eqref{eq:mpc_formulation} and assembles them into the form shown in \eqref{eq:kkt_system}.
A scaling step is performed with the Ruiz equilibration method $f_{\mathrm{Ruiz}}$, which improves the conditioning of the KKT matrix $\mathbf{A}_K$~\cite{ruiz2001scaling}.
The KKT matrix is then factorized (\eqref{eq:ldl}) with the \texttt{cuDSS} library, which will be discussed further in \autoref{sec:mpc_parallelization}~\cite{cudss}.
The factorized components of the KKT matrix are used to compute the QP iterations with repeated linear solves and updates with $f_{\mathrm{ADMM}}$ (\eqref{eq:backsolve_1}-\eqref{eq:backsolve_2}).

After $N_{\mathrm{QP}}$ iterations, the solution $\Delta \zvec_{N_{\mathrm{QP}}}$ to the QP is added to the initial guess $\zvec_{\mathrm{init}}[\cdot]$ with an optional line search (we simply keep $\alpha = 1$).
Finally, the spatial recursive Newton-Euler algorithm (RNEA)~\cite{featherstone2000robot} is used to compute feedforward torques from the first timestep of the MPC solution, $\qvec_{\mathrm{MPC}} = \qvec^*[0]$, $\qdvec_{\mathrm{MPC}} = \qdvec^*[0]$, and $\Fvec_{\mathrm{MPC}} = \Fvec^*[0]$.
The inverse dynamics $f_{\mathrm{RNEA}}$ are computed as
\begin{equation}
  \begin{aligned}
    \boldsymbol{\tau}_{\mathrm{RNEA}} & =
        \mathbf{M}(\qvec_{\mathrm{MPC}})\qddvec_{\mathrm{MPC}} + \mathbf{h}(\qvec_{\mathrm{MPC}}, \qdvec_{\mathrm{MPC}}) \\
        & - \mathbf{J}(\qvec_{\mathrm{MPC}})^\intercal \Fvec_{\mathrm{MPC}}.
    \label{eq:rnea}
  \end{aligned}
\end{equation}
The commanded joint torques for the MPC is then sent to the platform as
\begin{align}
    \boldsymbol{\tau}_{\mathrm{MPC}} =
        \mathbf{K}_p(\qvec_{\mathrm{MPC}} - \qvec) + \mathbf{K}_d(\qdvec_{\mathrm{MPC}} - \qdvec) + \boldsymbol{\tau}_{\mathrm{RNEA}}.
    \label{eq:mpc_torque}
\end{align}
A summary of the necessary computations is shown in Alg.~\ref{alg:gpu_mpc}.

\begin{algorithm}
\caption{GPU-Parallelized MPC via ADMM}\label{alg:gpu_mpc}
\begin{algorithmic}[1]
\State \textbf{in parallel:}
\State $\zvec_{\mathrm{init}}[\cdot] \gets {\color{codegreen}{f_{\mathrm{init}}}}(\qvec, \qdvec, \boldsymbol{\phi})$ \Comment{Initial guess}
    \State $\boldsymbol{\theta}_{\mathrm{MPC}} \gets {\color{codegreen}f_{\mathrm{param}}}(\qvec, \qdvec, \mathbf{c}, \mathbf{\boldsymbol{\phi}})$ \Comment{MPC parameter update}
    \State $\mathbf{A}_{K}, \mathbf{b}_K \gets {\color{codegreen}f_{\mathrm{KKT}}}(\zvec_{t}[\cdot], \boldsymbol{\theta}_t)$ \Comment{Form KKT system}
    \State $\hat{\mathbf{A}}_{K}, \hat{\mathbf{b}}_{K} \gets {\color{codegreen}f_{\mathrm{Ruiz}}}(\mathbf{A}_{K}, \mathbf{b}_K)$ \Comment{Ruiz equilibration}
    \State $\mathbf{L}_{K}, \mathbf{D}_{K} \gets $ \texttt{cuDSS}$_{\mathrm{fac}}(\hat{\mathbf{A}}_{K})$ \Comment{Eq. \eqref{eq:ldl}}
    \For{$i \gets 0$ to $N_{\mathrm{QP}}$} \Comment{QP iterations}
        \State $\vvec_{K} \gets $ \texttt{cuDSS}$_{\mathrm{solve}}(\hat{\mathbf{b}}_{K}, \mathbf{L}_{K}, \mathbf{D}_{K})$ \Comment{Eq. \eqref{eq:backsolve_2}}
        \State $\Delta \zvec_i, \hat{\mathbf{b}}_K \gets {\color{codegreen}f_{\mathrm{ADMM}}}(\vvec_K)$
    \EndFor
    \State $\zvec^*[\cdot] \gets \zvec[\cdot] + \alpha \Delta \zvec_{N_{QP}}$ \Comment{Line search}
    \State $\boldsymbol{\tau}_{RNEA}, \qvec_{MPC}, \qdvec_{MPC} \gets {\color{codegreen}f_{\mathrm{RNEA}}}(\zvec^*[\cdot])$
    \State return $\boldsymbol{\tau}_{MPC}, \qvec_{MPC}, \qdvec_{MPC}, \zvec^*[\cdot]$
\end{algorithmic}
\end{algorithm}

\subsection{Solver Parallelization}
\label{sec:mpc_parallelization}
To compute the MPC solution in parallel, we use the \texttt{CusADi} library~\cite{jeon2024cusadi}, which is a code-generation framework for generating parallelized GPU kernels.

As discussed in~\cite{jeon2024cusadi}, \texttt{CasADi} includes native symbolic LDL factorization and solve routines that could be used to address \eqref{eq:ldl} and \eqref{eq:backsolve_2}~\cite{Andersson2019_casadi}.
However, the code-generated kernel for factorizing the KKT matrices requires millions of instructions, nearing CPU memory limits for compilation.
Instead, we use NVIDIA's \textbf{D}irect \textbf{S}parse \textbf{S}olve (\texttt{cuDSS}) library to address the factorization and linear solve steps, which is optimized for parallel GPU computation.
The library contains routines for sparse LDL factorization of indefinite matrices and linear solves, suited for the KKT system in \eqref{eq:kkt_system} \cite{cudss}.
The expensive factorization step only needs to be computed once per MPC solve, while the inexpensive linear solves can reuse the factorized matrices to solve for the primal and dual variables in parallel.

The remaining auxiliary functions, highlighted in green in Alg.~\ref{alg:gpu_mpc}, are generated for the GPU with \texttt{CusADi} and are executed in parallel for each MPC solve.
The number of instructions range between $10^3$ to $10^6$ for these functions, and are straightforward to generate and compile.
\section{Residual Architecture}

We investigate the use of residual policy learning to improve and adjust the outputs of an MPC locomotion controller for \mithumanoid.
As shown in \autoref{fig:diagram}, we evaluate the MPC and residual policy at the same control frequency of 100 Hz for both training and deployment.
The MPC is parallelized on the GPU for training, and deployed on the CPU with OSQP for online evaluation.
The residual network is a simple multilayer perceptron (MLP) with three layers, each with 256 exponential linear unit (ELU) nodes.
The policy observes the full state of the robot, contact phase variables, and the value from the MPC, denoted as
$\mathbf{o} = [\mathbf{p}^\intercal \;
               \boldsymbol{\theta}^\intercal \;
               \qvec_j^\intercal \;
               \boldsymbol{\omega}^\intercal \;
               \vvec^\intercal \;
               \qdvec_j^\intercal \;
               \boldsymbol{\phi}^\intercal \;
               V_\mathrm{MPC}
               ]^\intercal \in \mathbb{R}^{54}
$, and outputs actions for the legs $\mathbf{a} \in \mathbb{R}^{10}$.
To reduce reward shaping complexity and tuning, we chose to limit our scope to adapting the leg torques only for locomotion.
Although the residual networks were tested with arm actions as well, we observed only minimal deviations from the MPC controller's predictions for the upper body.
Additionally, simple box constraints on the arm joints were sufficient to prevent self-collision from the arms through the MPC horizon.
A more thorough treatment on learning adaptive arm motion is given in~\cite{lee2025learning}.

Surprisingly, the residual policy is still able to learn and reliably adjust to the MPC controller despite being given very limited information about its outputs.
While we originally allowed the network to observe the output torques and predicted states, we found that this made sim-to-sim transfer and sim-to-real transfer less reliable, as the distributions of these quantities were sensitive to the IsaacGym simulator.

\subsection{Reward Design}
For this work, we study the role of the baseline MPC controller as a \textit{prior} for guiding the policy towards favorable states, rather than as an expert actor to be imitated.
Consequently, we limit the scope of our experiments to "standard" RL rewards given directly from environment transitions to clearly isolate the effects of the residual architecture\footnote{As opposed to adversarial or distribution-imitating frameworks~\cite{ho2016generative,tang2024humanmimic}. Future work could investigate the effect of other reward paradigms attached to a residual architecture.}.
We wish to demonstrate that the residual policy allows the humanoid to learn behaviors beyond the capabilities of the designed MPC controller without overriding its outputs entirely.
To this end, we design a set of reward functions that are generally aligned with the MPC objectives, but include nondifferentiable and sparse terms that would be difficult to directly optimize for in an online setting, such as self-collision avoidance or termination conditions.
We give self-collision penalties when contact forces are detected between the humanoid's links, and termination conditions when the floating base has large deviations in velocities, orientation, or height, as in~\cite{jeon2023benchmarking}.
A full list of the rewards used is detailed in \autoref{tab:rewards}.

\begin{table}[tbp]
\centering
\caption{Reward functions and their weights.}
\label{tab:rewards}
\begin{tabular}{lcl}
\toprule
\textbf{Reward Name} & \textbf{Weight} & \textbf{Function} \\
\midrule
Lin. vel. tracking & 10.0 & $\exp\left(- \norm{\frac{{\mathbf{c}_{x,y} - \mathbf{v}_{x,y}}}{1 + |\mathbf{c}_{x,y}|} }_2^2 / \sigma \right)$ \\
Ang. vel. tracking & 5.0 & $\exp\left(- \norm{\mathbf{c}_{\omega} - \boldsymbol{\omega}_z}_2^2 / \sigma \right)$ \\
1$^\mathrm{st}$ order action rate & -1e-3 & $\norm{ (\mathbf{a}_t - \mathbf{a}_{t-1})/\Delta t }_2^2 $ \\
2$^\mathrm{nd}$ order action rate & -1e-4 & $\norm{ (\mathbf{a}_t - 2\mathbf{a}_{t-1} + \mathbf{a}_{t-2})/\Delta t  }_2^2 $ \\
Torques & -1e-4 & $\norm{\boldsymbol{\tau}}_2^2$ \\
Orientation & 1.0 & $\exp\left(- \norm{\mathbf{g}_{x,y}}_2^2 / \sigma \right)$ \\
Height & 1.0 & $\exp\left(- \norm{\mathbf{c}_{z} - \mathbf{p}_z}_2^2 / \sigma \right)$ \\ 
Joint regularization & 1.0 & $\frac{1}{n_j}\sum_{j=1}^{n_j} \norm{\mathbf{q}_j - \hat{\mathbf{q}}_j}_2^2$\\
Self-collision & -1.0 & $\mathbbm{1}_{\mathrm{collision}}$ \\
Termination & -100 & $\mathbbm{1}_{\mathrm{terminate}}$ \\
\bottomrule
\end{tabular}
\end{table}

The set of rewards given to the system is intentionally minimal.
We avoid giving overly specific rewards such as foot guidance, air time, or contact-scheduling terms to study how the MPC can guide policy training towards favorable states with minimal tuning, as detailed in \autoref{sec:training_guidance}.
More importantly, we seek to move away from fitting complex rewards to desirable locomotion behavior in favor of using physically-grounded models that can plan and adapt behavior online.

\subsection{Blending Strategy}
\label{sec:blending}
It is important that the residual policy has little effect on the baseline controller when initialized~\cite{johannink2019residual,silver2018residual,luo2024residual}.
This is to ensure that the residual controller is aligned with the control prior at the start of training, instead of destabilizing it with random outputs.
If the action space of the policy output is the same as that of the nominal policy, this can easily be achieved by initializing the policy network to have weights and biases of zero (e.g., a position controlled manipulator, with position-space actions for the residual policy).
However in our case, our baseline MPC outputs torques, while policies are typically trained with joint setpoint actions.
To study the effect of the action space on the residual architecture, we consider three different strategies for blending these outputs:
\begin{enumerate}
  \item \textbf{Joint action, joint blending:}
  \begin{equation} \label{eq:joint-joint}
    \boldsymbol{\tau} = \mathbf{K}_p(\qvec_{\mathrm{cmd}} + \lambda \mathbf{a} - \qvec) + \mathbf{K}_d(\qdvec_{\mathrm{cmd}} - \qdvec) + \boldsymbol{\tau}_{\mathrm{cmd}},
  \end{equation}
  \item \textbf{Joint action, torque blending:}
  \begin{align} \label{eq:joint-torque}
    \boldsymbol{\tau}_{\mathrm{residual}} &= \mathbf{K}_p (\mathbf{a} + \hat{\qvec} - \qvec) - \mathbf{K}_d \mathbf{\dot{q}} \\
    \boldsymbol{\tau} &= \boldsymbol{\tau}_{\mathrm{MPC}} + \lambda \boldsymbol{\tau}_{\mathrm{residual}},
  \end{align} 
  \item \textbf{Torque action, torque blending:}
  \begin{equation} \label{eq:torque-torque}
    \boldsymbol{\tau} = \boldsymbol{\tau}_{\mathrm{MPC}} + \lambda \mathbf{a},
  \end{equation}
\end{enumerate}
where $\mathbf{a}$ is the action output from the residual policy, $\hat{\qvec}$ are default joint positions, and $\boldsymbol{\lambda}$ is a scalar blending factor.

In the first case, the policy modifies the joint setpoint command from the MPC torques computed in \eqref{eq:mpc_torque}.
Here, a zero-initialized policy will not affect the MPC outputs, but the policy is relative to $\qvec_{\mathrm{cmd}}$ which is non-stationary.
In the second case, the policy outputs joint setpoints relative to default joint positions, and the final torques from the MPC and residual policy are blended together.
However, the residual policy will have non-zero torques even if the network is zero-initialized.
Consequently, we introduce a weighting parameter to control the initial influence of the policy.
In the third case, the policy directly outputs torques, which are then combined with the MPC torques.
We analyze how the choice of action space affects the residual policy's ability to learn from the MPC prior in \autoref{sec:training_guidance}

\subsection{Training Procedure}
\label{sec:training}
We use the IsaacGym simulation environment to gather simulation rollouts of our residual controller~\cite{makoviychuk2021isaac}.
Due to GPU VRAM limitations from the parallelized MPC, we limit the number of parallel environments to 2048.
To train the policy, we use Proximal Policy Optimization (PPO)~\cite{schulman2017proximal} with a clipped objective and generalized advantage estimation (GAE)~\cite{schulman2015high}.
The same PPO hyperparameters from~\cite{jeon2023benchmarking} are used for all experiments, and a summary of the training procedure is shown in Alg.~\ref{alg:residual_training}.
Note that any RL algorithm could be used to optimize the policy, but we choose PPO for its simplicity and effectiveness in continuous control tasks.

\begin{algorithm}
\caption{Residual Policy Training}\label{alg:residual_training}
\begin{algorithmic}[1]
\Require Initial policy $\pi_{\theta}$, $\pi_{\mathrm{MPC}}$
\For{$n=0, \dots, N-1$ episodes}
    \State Initialize exporation noise $\mathcal{N}$
    \State Sample initial state $\mathbf{s}_0 \sim \mathcal{D}$
    \For{$t=0, \dots, H-1$ timesteps}
        \State Compute MPC: $\zvec^*_t[\cdot], \boldsymbol{\tau}_{\mathrm{MPC}, t} \leftarrow$ Alg. \ref{alg:gpu_mpc}
        \State Sample policy: $\mathbf{a}_t \sim \pi_{\theta}(\mathbf{s}_t, \zvec^*_t[\cdot]) + \mathcal{N}_t$
        \State Blend outputs: $\boldsymbol{\tau}_t \leftarrow f_{\mathrm{blend}}(\boldsymbol{\tau}_{\mathrm{MPC}, t}, \mathbf{a}_t)$ \Comment{\ref{sec:blending}}
        \State Simulate envs.: $\mathbf{s}_{t+1} \sim \mathcal{T}(\mathbf{s}_t, \boldsymbol{\tau}_t)$
        \State Store transitions: $\mathcal{R} \leftarrow (\mathbf{s}_t, \boldsymbol{\tau}_t, \mathbf{s}_{t+1})$
    \EndFor
    \State Sample transitions: $(\mathbf{s}, \boldsymbol{\tau}, \mathbf{s}) \sim \mathcal{R}.$
    \State Optimize policy with RL: $\pi_{\theta_{t+1}}$. \Comment{PPO}
\EndFor
\end{algorithmic}
\end{algorithm}

\section{Results}
\label{sec:results}
We study the performance of the residual architecture for learning locomotion tasks on~\mithumanoid.
All reported results are collected on a desktop computer equipped with an NVIDIA RTX 3090 GPU and a 10th Gen. Intel Core i9-10900K CPU.
Ablation studies and various sweeps are evaluated on an NVIDIA A100 GPU.

We consider three controllers for comparison.
\begin{itemize}
    \item \textbf{MPC}: The fixed MPC controller described in \autoref{sec:mpc_formulation}.
    \item \textbf{Residual}: The proposed residual architecture.
    \item \textbf{End-to-End}: A controller learned with conventional RL end-to-end.
\end{itemize}
Unless specified otherwise, each of these controllers is given the same rewards, range of commands, initial configurations, disturbances, domain randomization, etc. during training to ensure a fair comparison.

To clearly analyze the effect of incorporating model-based priors into RL, we limit our scope to these three controllers.
While many state of the art controllers exist for humanoid platforms (both learned and model-based), we wish to isolate the relative benefits over strictly model-based and strictly learning-based benchmarks, which could transfer to more complex training/controller architectures.
Therefore, we limit our scope to studying the advantages of the residual architecture for combining MPC and RL specifically, rather than benchmarking the absolute performance of the residual policy against all other possible humanoid controllers.


\subsection{GPU-Parallelization of MPC}
\label{sec:results_parallelization}
\begin{figure}[tbp] 
    \centering
    \includegraphics[width=\columnwidth]{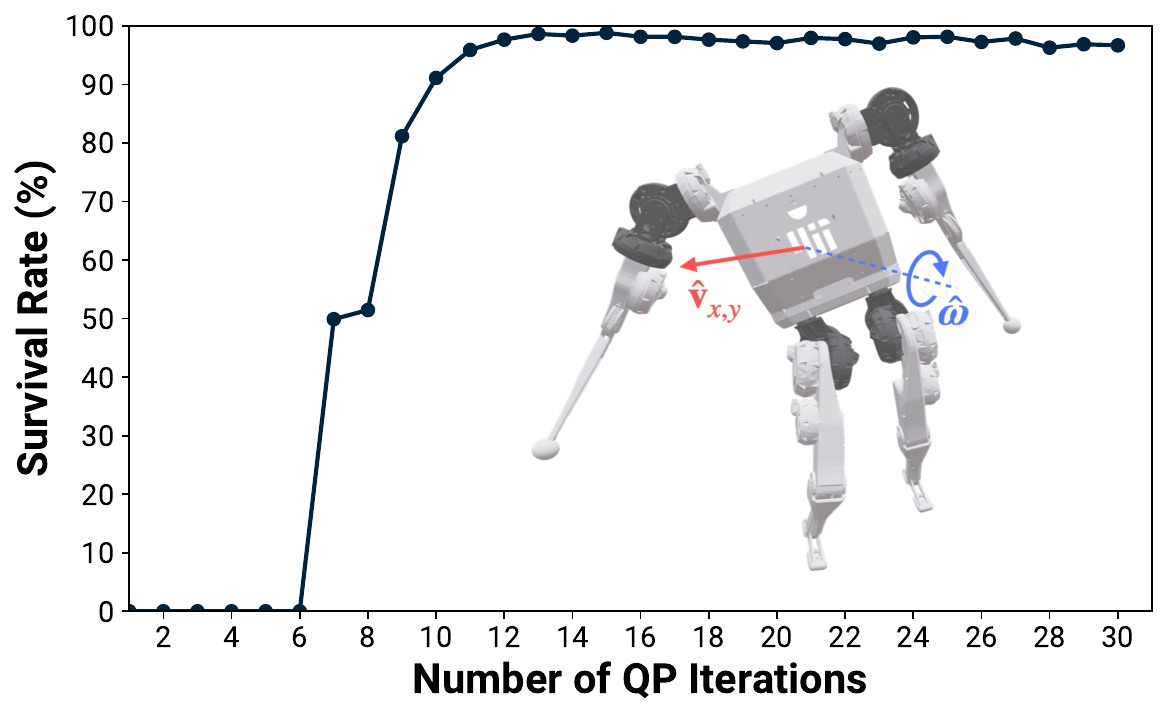}
    \caption{
        Plot of survival rate vs. number of QP iterations for MPC environments subject to randomized initial velocities in the range $\norm{\hat{\vvec}_{x, y}} \leq 0.5$ m/s and $\norm{\hat{\boldsymbol{\omega}}} \leq 0.5$ rad/s.
        Beyond a certain number of iterations, the marginal benefit is negligble while increasing computation time.
    }
    \label{fig:qp_iterations}
\end{figure}

We begin by characterizing the computations necessary for the MPC controller outlined in \autoref{sec:mpc_formulation}.
From Alg. \ref{alg:gpu_mpc}, the only way to reduce the total number of computations is by reducing $N_{\mathrm{QP}}$ - all other calculations are strictly necessary for evaluation.
Consequently, we sweep the controller across a range of initial linear and angular base velocities while stepping with zero velocity command, and record the percentage of robots that survive termination (defined with simple bounds on height and self-collision within five seconds of the initial disturbance).
As shown in \autoref{fig:qp_iterations}, we find that the marginal benefit of additional QP iterations decreases significantly once the solution is "good enough" under these simple conditions.
Conservatively, we set $N_{\mathrm{QP}} = 25$ for the controller for the remainder of the work.
During hardware deployment on the CPU, this can be evaluated between 200-300 Hz single-threaded.

To generate the parallelized controller, the \texttt{CasADi} objects for each of the functions in Alg. \ref{alg:gpu_mpc} must be unrolled and compiled as a \texttt{CUDA} kernel.
This can be a significant bottleneck, as \texttt{CasADi} functions can easily contain millions of instructions, equating to a kernel with millions of lines of code to compile.
By making use of \texttt{cuDSS}, we avoid compiling kernels for the KKT factorization and backsolve, the most computationally expensive steps.
\autoref{tab:casadi_functions} presents details about the functions that were parallelized with \texttt{CusADi}.
The most expensive functions to generate and evaluate are those that form the KKT matrices - namely, this consists of the cost Hessians and constraint Jacobians of the MPC formulation. 

\begin{table}[tbp]
    \centering
    \caption{Functions for MPC parallelization ($N_{\mathrm{QP}} = 25$).}
    \label{tab:casadi_functions}
    \renewcommand{\arraystretch}{1.2}
    \begin{tabular}{lccc}
        \hline
        \textbf{Function} & \textbf{Num. instr.} & \textbf{Compile time (s)} & \textbf{Eval. time (ms)} \\
        \hline
        ${\color{codegreen}{f_{\mathrm{init}}}}$    & 9.64e2 & 36.6 & 2.35 $\pm$ 0.1 \\
        ${\color{codegreen}f_{\mathrm{param}}}$     & 9.19e2 & 38.5 & 8.03 $\pm$ 0.4 \\
        ${\color{codegreen}f_{\mathrm{KKT}}}$       & 1.05e6 & 1.91e3 & 27.46 $\pm$ 0.9 \\
        ${\color{codegreen}f_{\mathrm{Ruiz}}}$      & 3.72e5 & 1.53e3 & 23.65 $\pm$ 0.8 \\
        ${\color{codegreen}f_{\mathrm{ADMM}}}$      & 2.45e4 & 133.1 & 3.11 $\pm$ 0.2 \\
        ${\color{codegreen}f_{\mathrm{RNEA}}}$      & 7.34e3 & 39.2 & 0.27 $\pm$ 0.1 \\
        \texttt{cuDSS}$_{\mathrm{fac}}$             & -      & - & 81.75 $\pm$ 1.7 \\
        \texttt{cuDSS}$_{\mathrm{solve}}$           & -      & - & 12.72 $\pm$ 0.4 \\
        \hline
        \textbf{Total}      & -      & - & 492.82 $\pm$ 2.8 \\
        \hline
    \end{tabular}
\end{table}

\begin{figure}[tbp]
    \centering
    \includegraphics[width=\columnwidth]{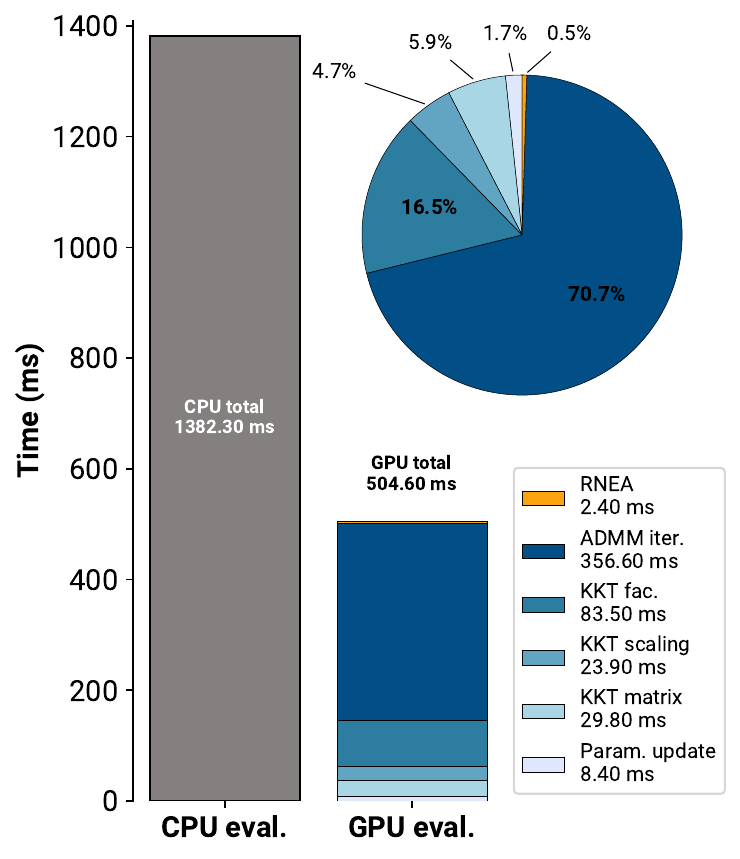}
    \caption{
        Computation time required for evaluating the full MPC controller on the CPU and GPU across 1000 environments in IsaacGym.
        We optimistically estimate the required time for the CPU assuming 10 perfectly parallel cores, and do not include the time necessary for CPU-GPU data transfer.
        More detailed benchmarking results for \texttt{CusADi} on the CPU and GPU are shown in~\cite{jeon2024cusadi}.
    }
    \label{fig:gpu_times}
\end{figure}

Next, we study the computational load incurred from the GPU-parallelized MPC.
In \autoref{fig:gpu_times}, the speedup relative to CPU evaluation is shown.
For the CPU evaluation time, we optimistically estimate perfect parallelization between the ten cores of the i9-10900K Intel CPU, and do not include the additional time necessary to transfer the data from the CPU to the GPU for training, which is the primary bottleneck~\cite{jeon2024cusadi}.
Despite this, GPU evaluation is still roughly 2.5x faster than CPU evaluation for parallelization across 1,000 environments.
For each simulation step, the majority of the evaluation time is spent performing ADMM iterations (70.7\%).
The total evaluation time could be further reduced by adjusting the $N_{\mathrm{QP}}$ parameter to be less conservative.

While the computations are efficiently parallelized with \texttt{CusADi}, solving thousands of optimization problems at each timestep of the simulation still incurs significant overhead compared to standard RL training.
While the authors of \cite{jenelten2024dtc} evaluate their MPC formulation at 2-3 Hz, we evaluate MPC at the same frequency as the policy (100 Hz) which greatly increases total training time.
For 1,000 environments, a single simulation step takes roughly 0.5 seconds, as shown in \autoref{fig:gpu_times}.
This scales linearly with the number of simulation steps performed per PPO iteration.
For our work, we use 24 steps, corresponding to roughly 12 seconds per gradient update.
This is approximately 8x slower than end-to-end RL on the same computer.
However, we show in \autoref{sec:training_guidance} that the parallelized MPC prior significantly shapes the convergence of the policy, steering exploration towards more desirable behavior.
Although training requires more time, the final policy can be more reliably guided than by painstakingly tweaking reward weights and environment parameters.

\begin{figure}[tbp] 
    \centering
    \includegraphics[width=\columnwidth]{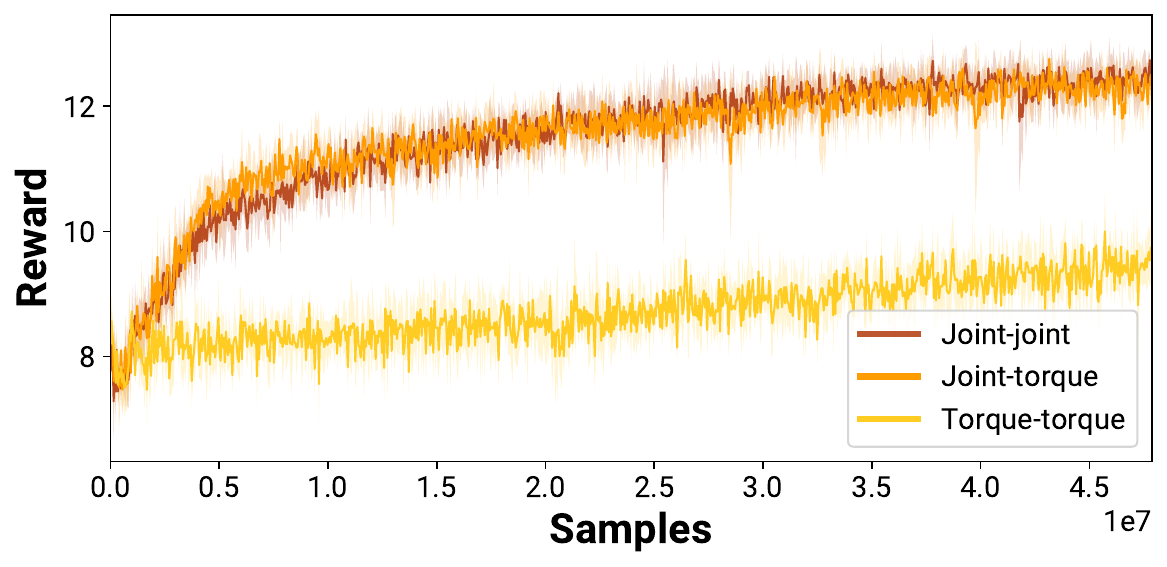}
    \caption{
        Reward comparison of three different blending strategies between MPC and the residual policy.
        While there is little difference between joint-space action representations for the residual policy, a torque-space action representation performs significantly worse.
    }
    \label{fig:action_space_comparison}
\end{figure}
\begin{figure}[tbp] 
    \centering
    \includegraphics[width=\columnwidth]{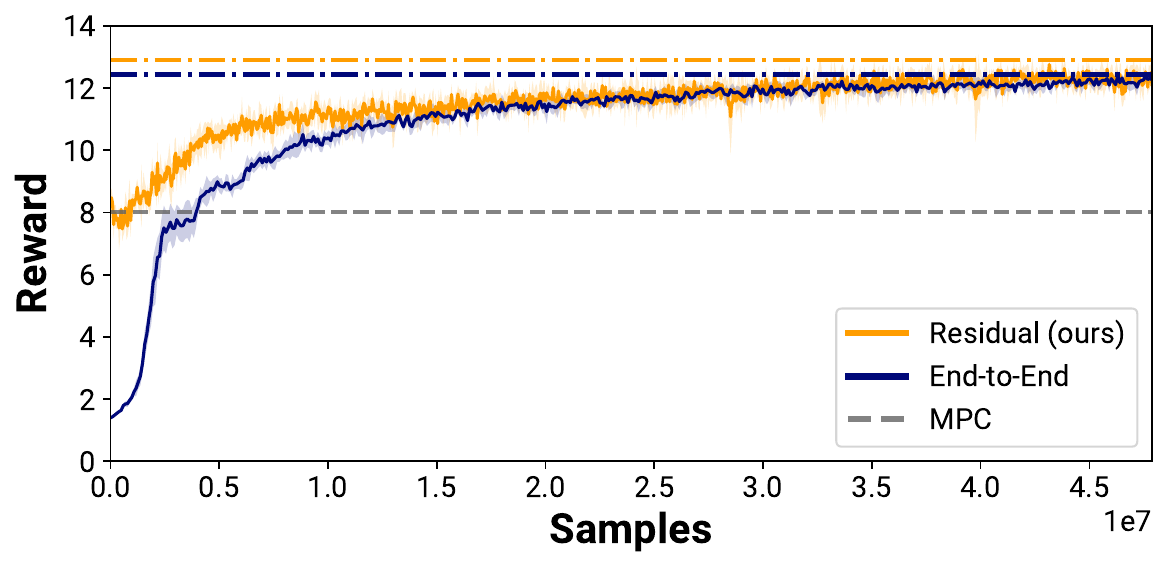}
    \caption{
        The reward incurred during training between the residual and end-to-end controllers.
        The fixed MPC prior maintains a constant value as it contains no network to be optimized from PPO iterations.
        The residual architectures shows improvements in sample efficiency and asymptotic performance compared to the end-to-end training.
    }
    \label{fig:reward_comparison}
\end{figure}
\begin{figure*}[tbp]
  \centering
  \includegraphics[width=0.9\textwidth]{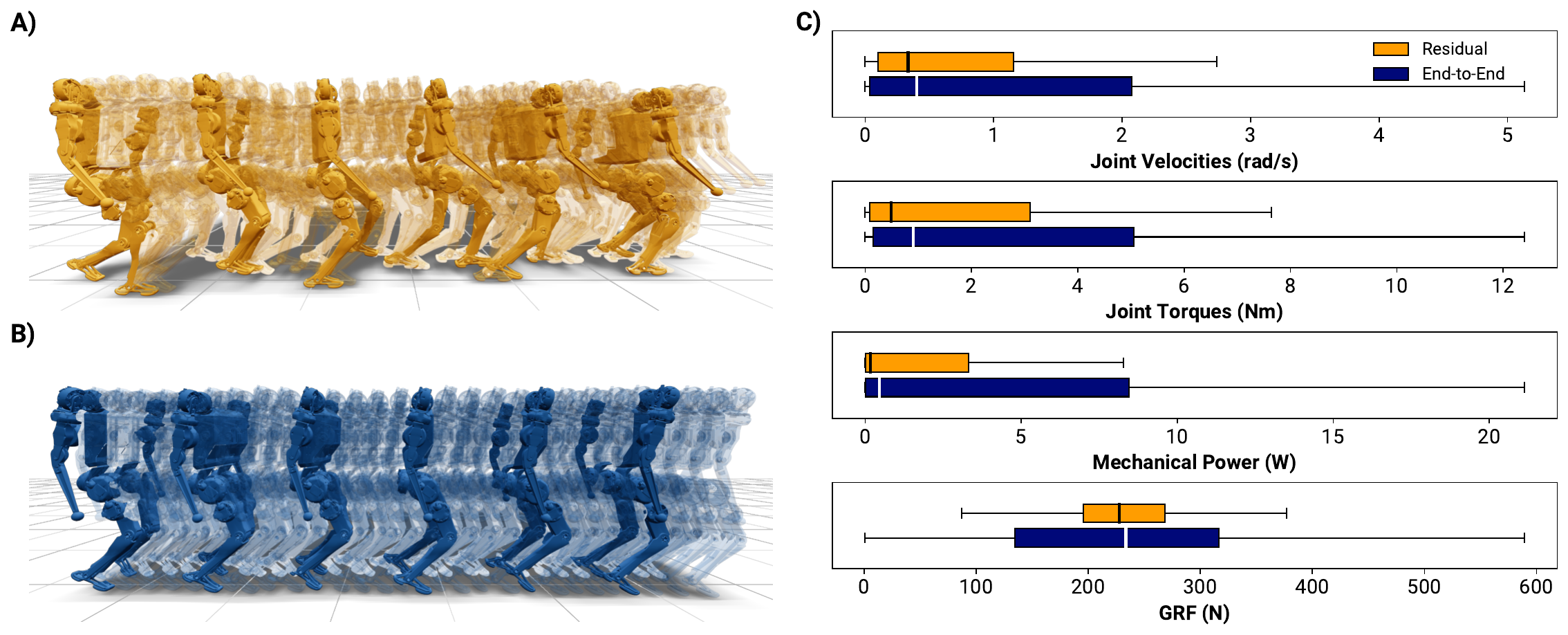}
  \caption{
        \textbf{A)} Snapshots of the residual policy in motion walking forward at 1 m/s.
        \textbf{B)} Snapshots of the end-to-end policy trained with the same set of rewards.
        The policy exhibits unrealistic "gliding" motion, where the feet rapidly tap the ground to traverse the environment.
        \textbf{C)} Comparative distributions between the residual and end-to-end policies.
        Across all metrics, the residual policy exhibits lower medians and spreads compared to the end-to-end policy, suggesting more viable sim-to-real transfer.
    }
  \label{fig:policy_state_space_comparison}
\end{figure*}
\begin{figure}[btp] 
    \centering
    \includegraphics[width=0.9\columnwidth]{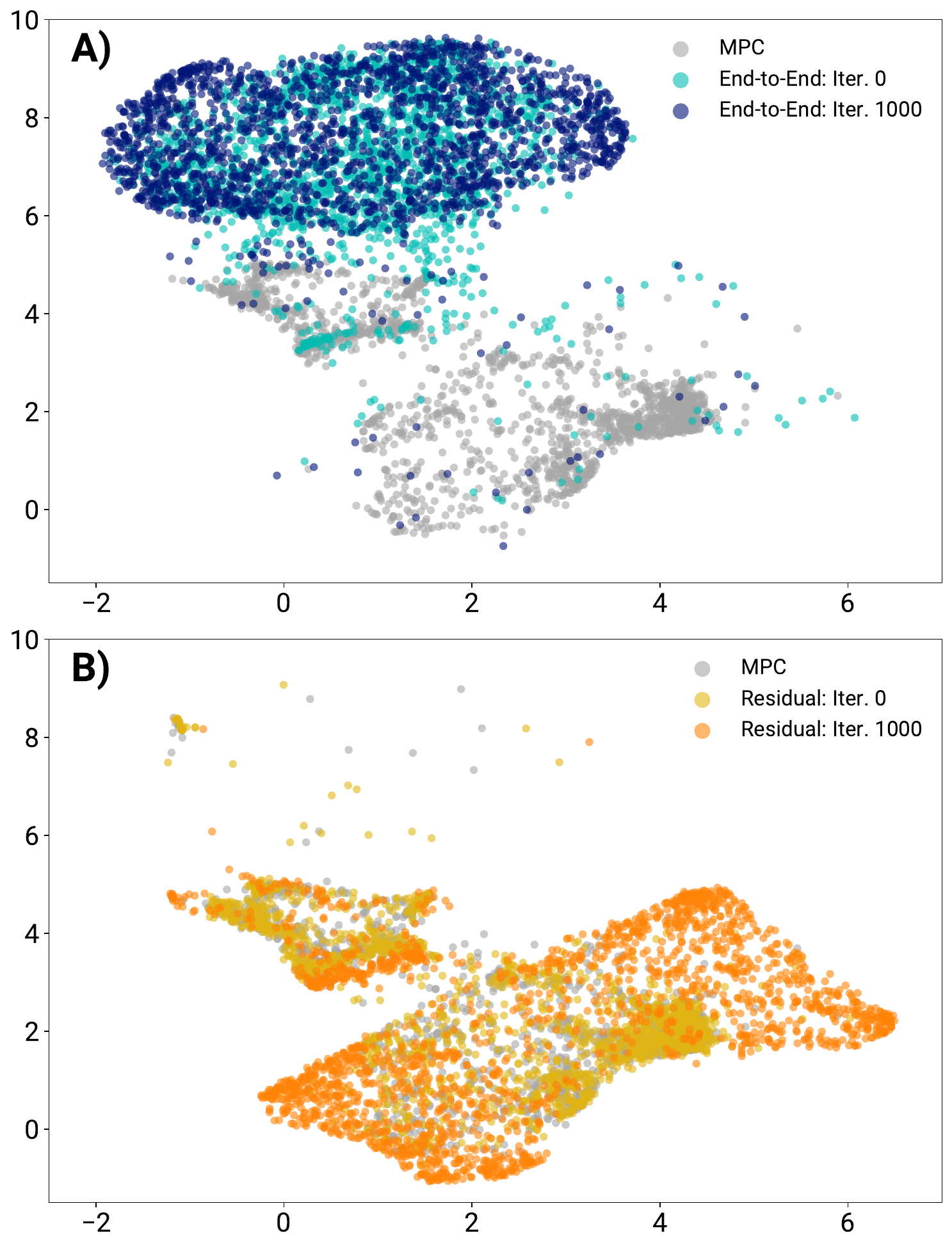}
    \caption{
        Distribution of states experienced at the start of training and at training convergence, visualized with UMAP dimensionality reduction.
        The states incurred from the fixed MPC controller are shown in gray.
        \textbf{A)} State visualization for the end-to-end policy. 
        \textbf{B)} State visualization for the residual policy.
    }
    \label{fig:umap_training_state_dist}
\end{figure}
\begin{figure*}[tbp]
  \centering
  \includegraphics[width=0.9\textwidth]{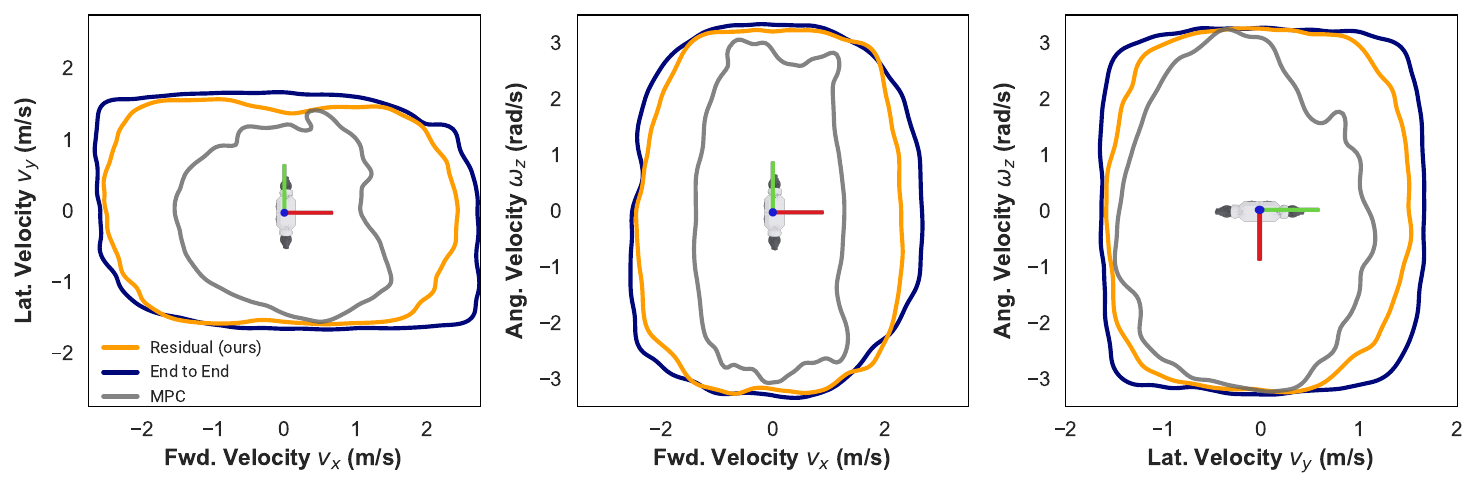}
  \caption{
    95\% kernel density estimate boundaries of the trackable velocities for each policy.
    The residual policy network substantially improves the viable range of commands over the MPC baseline.
  }
  \label{fig:velocity_boundaries}
\end{figure*}
\begin{figure}[tbp] 
    \centering
    \includegraphics[width=0.95\columnwidth]{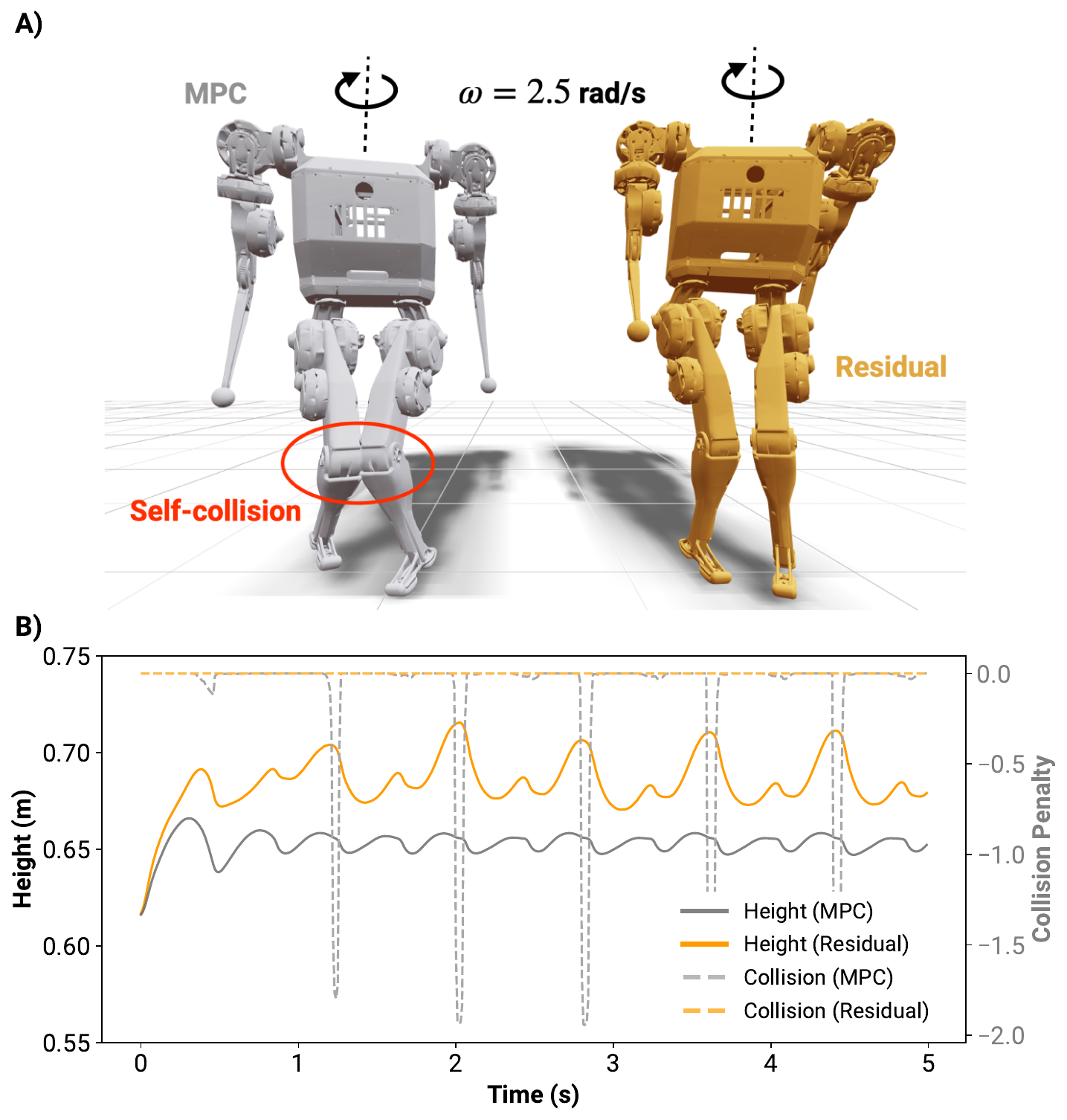}
    \caption{
        \textbf{A)} The MPC policy collides at the knees when turning at 2.5 rad/s, while the residual does not.
        \textbf{B)} Plot of the controllers' height and self-collision penalty while turning.
    }
    \label{fig:collision_avoidance}
\end{figure}
\begin{figure*}[tbp]
  \centering
  \includegraphics[width=\textwidth]{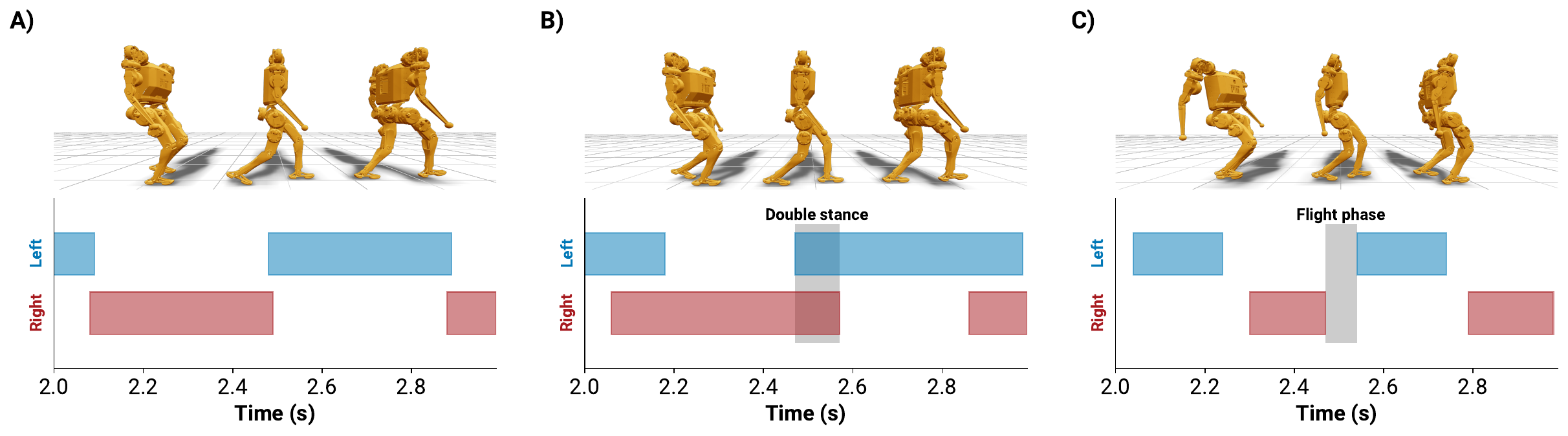}
  \caption{
    The gait parameters of the MPC are varied after training the residual network to induce double stance and flight phase in the contact schedule.
    While the MPC is incapable of maintaining these gaits stably, the residual network can adapt to these out of distribution contact modes despite not experiencing them during training.
  }
  \label{fig:contact_scheduling_study}
\end{figure*}
\begin{figure*}[tbp]
  \centering
  \includegraphics[width=\textwidth]{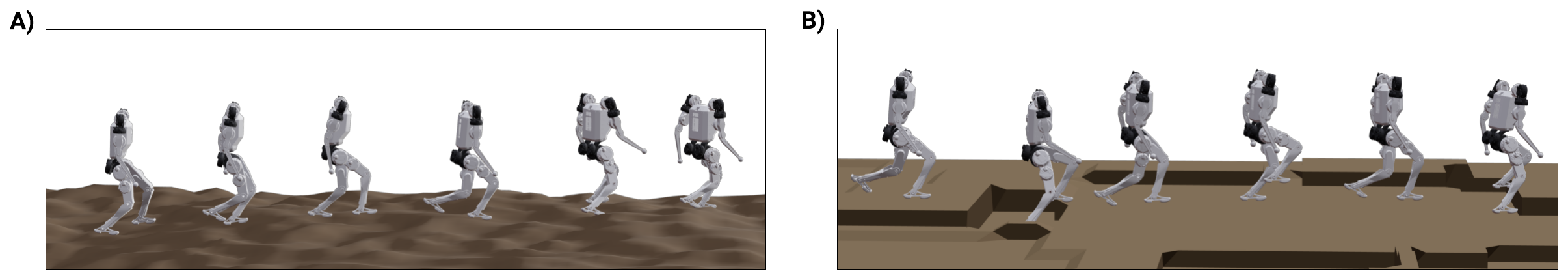}
  \caption{
        The residual policy is capable of traversing terrain never seen during training.
        Despite only being trained over flat ground, the residual is capable of navigating various surfaces including \textbf{A)} uneven, rough terrain and \textbf{B)} discrete, sloped terrain.
  }
  \label{fig:terrain}
\end{figure*}
\begin{figure*}[tbp]
  \centering
  \includegraphics[width=\textwidth]{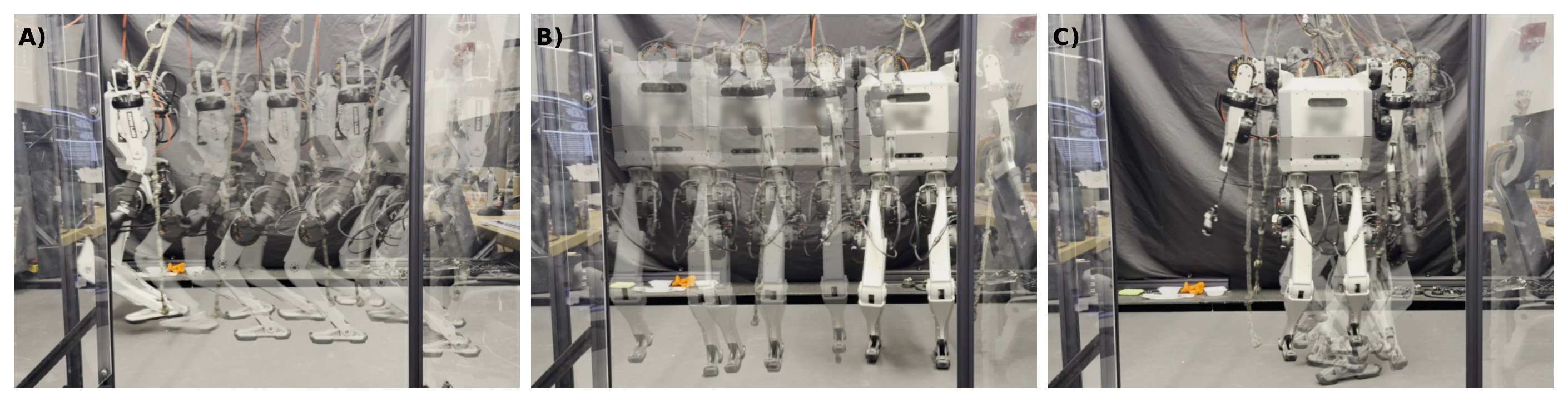}
  \caption{
    Hardware validation of the residual policy architecture on~\mithumanoid, for \textbf{A)} forward walking, \textbf{B)} lateral walking, and \textbf{C)} turning.
  }
  \label{fig:hardware}
\end{figure*}

\subsection{Training}
\label{sec:training_guidance}

With the fixed set of rewards from \autoref{tab:rewards}, we first consider the effect of the proposed blending strategies.
Specifically, we vary the action space of the residual policy (joint setpoint/torque) and the manner of blending (joint-space/torque-space).
We train each of these residual policies with the same set of rewards and a zero-initialized policy network, as described in \cite{silver2018residual}, and set the blending factor $\lambda = 0.1$.

As shown in \autoref{fig:action_space_comparison}, we find that the residual policy learns more effectively from a joint-space representation of its actions.
Although the \textit{joint-torque} training produces nonzero torque at the onset of training as in \autoref{eq:joint-torque}, the difference to the \textit{joint-joint} training is negligible.
The torque space representation controls the joints at the acceleration level, making it more difficult for the policy to smoothly explore actions with high advantage.
Additionally, the scale of torque-space actions are significantly larger than joint-space actions.
With the same action and action rate regularization weights across the experiments, this could have affected the convergence of the \textit{torque-torque} training.

Ultimately, we use the \textit{joint-torque} strategy for blending the MPC and residual policy outputs.
While both the \textit{joint-joint} and \textit{joint-torque} experiments demonstrated similar performance, in the event of a diverging MPC solution, the \textit{joint-joint} strategy would output actions relative to potentially infeasible $\mathbf{q}_{\mathrm{MPC}}$ setpoints.
To avoid this potential failure mode, we use the \textit{joint-torque} strategy to ensure reasonable policy outputs regardless of the MPC solution. 

With the blending strategy established, we compare the reward curves for the residual policy against the MPC-only and RL-only baselines, shown in \autoref{fig:reward_comparison}.
While both the end-to-end and residual policies show significant improvements over the reward incurred by the MPC-only controller, the residual training converges faster and to a higher asymptote than the end-to-end training.
As discussed in \cite{silver2018residual, johannink2019residual}, the residual architecture allows the policy to leverage strong control priors by being initialized at the performance of the baseline MPC. 

While the residual training exhibits improvements in sample efficiency compared to the end-to-end training, the wall clock time is significantly longer.
For 1,000 iterations of PPO, end-to-end training only requires around 30 minutes, whereas the residual can range from three to four hours.
Purely from the perspective of time to convergence, evaluating the parallelized MPC in the RL loop is far more expensive than standard end-to-end training.

However, the qualitative and quantitative differences in the converged policies are notable.
As shown in \autoref{fig:policy_state_space_comparison}, the end-to-end and residual controllers exhibit entirely different behaviors for locomotion.
Despite being given the exact same rewards, environments, initialization conditions, and so on, the end-to-end policy learns to rapidly oscillate the ankle joint of the humanoid to "glide" across the floor, exploiting the physics of the environment.
While this behavior may be feasible in the IsaacGym simulation, it is unlikely that this would transfer safely to hardware.

In contrast, the residual policy learns to stay close to the MPC solutions, exhibiting clear stepping behavior at the same frequency as the MPC.
When we compare key characteristics between the two policies such as the joint velocities, torques, mechanical power, and vertical ground reaction force, we find that the residual policy has a smaller median and range overall.
Despite being given the same rewards related to actuation (torque and action rate penalization), the residual policy demonstrates lower extremes in the experienced joint velocities, torques, mechanical power expended, and ground reaction forces.
The inclusion of the MPC prior greatly affects policy training, such that even for the same set of rewards, the converged behavior ranges from physically unrealistic to viable for hardware deployment.

More broadly, RL can be sensitive to the reward signals, initialized states, curricula, and algorithm hyperparameters of the training setup.
Algorithms like PPO theoretically can converge to global optima, but in practice, it is difficult to balance exploration and exploitation of the policy to achieve desirable behavior.
The feedback loop of tuning a training procedure can also be time-consuming and unintuitive.
As additional reward shaping terms are added to guide the policy, it becomes more difficult to parse the individual and possibly combinatoric effects from all the rewards.

Instead of engineering complex rewards and/or curricula to guide the policy towards desirable states, control priors can serve to initialize the policy "close to" a desired behavior.
The residual architecture can be viewed as "warm start" in the search for an optimal policy, initialized as an MPC controller.

To demonstrate this, we investigate how the states explored by the residual architecture and the end-to-end policy differ through training.
We compare the states experienced from noisy, exploratory actions at the initial stage of training with the final stage of training for the two policies, alongside the distribution from the (fixed) MPC controller alone.
The states from 24 simulation steps are collected, the same amount of data seen from each iteration of PPO.
To visualize the 48 dimensional state-space of the humanoid platform, we use the Uniform Manifold Approximation and Projection (UMAP) technique~\cite{McInnes2018UMAP} to reduce the dimension of the state data to a 2D manifold, as shown in \autoref{fig:umap_training_state_dist}.
For the UMAP metric, we use the L-1 norm because of the differences in scale between state variables (e.g., joint angles vs. base linear velocities).
However, the shape of the mappings appear to be similar regardless of the metric used.

We find that despite being given the same reward signal, the convergence of the states is strongly affected by the bias from the MPC prior.
At the start of training, with a zero-initialized policy network, the state distribution of the residual policy is predictably close to that of the MPC controller.
This influence throughout training alters the evolution of the policy - whereas the end-to-end policy converges to an entirely different distribution of states, the residual remains close to the initial MPC distribution despite being given the same reward signal.
Some aspects of the UMAP visualization are similar, most likely corresponding to base velocities or initialization conditions, but overall, the differences in state distribution reflect the qualitative differences in locomotion behavior shown in \autoref{fig:policy_state_space_comparison}.
The MPC prior biases the visited states during training, and consequently, the converged locomotion behavior of the residual policy.
It acts as a "warm-start" for the RL training, initializing state-action pairs close to the MPC controller.



\begin{figure*}[tbp]
  \centering
  \includegraphics[width=\textwidth]{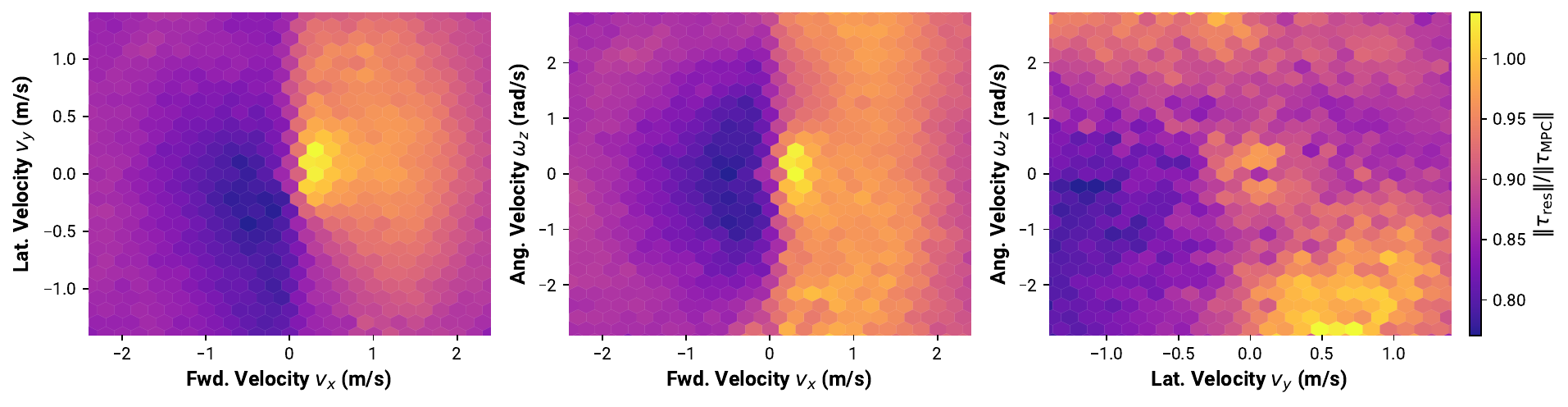}
  \caption{
    The ratio of residual to MPC torque magnitudes over the commanded velocities.
    The residual policy is leveraged asymmetrically, especially for $v_x$ commands.
  }
  \label{fig:residual_vs_command}
\end{figure*}

\begin{figure}[tbp] 
    \centering
    \includegraphics[width=0.9\columnwidth]{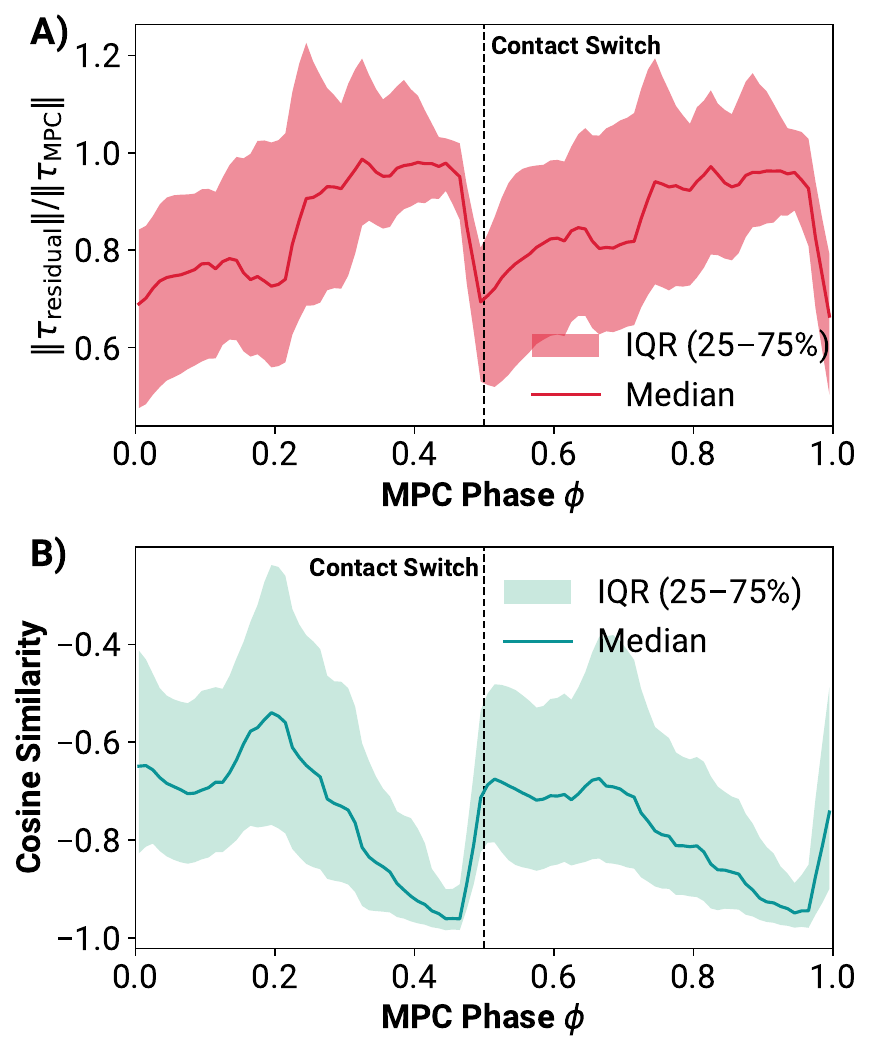}
    \caption{
        \textbf{A)} The percentage of torque from the residual with respect to MPC as a function of phase.
        \textbf{B)} The cosine similarity between the residual and MPC torques. 
    }
    \label{fig:residual_analysis}
\end{figure}

\begin{figure*}[tbp]
  \centering
  \includegraphics[width=\textwidth]{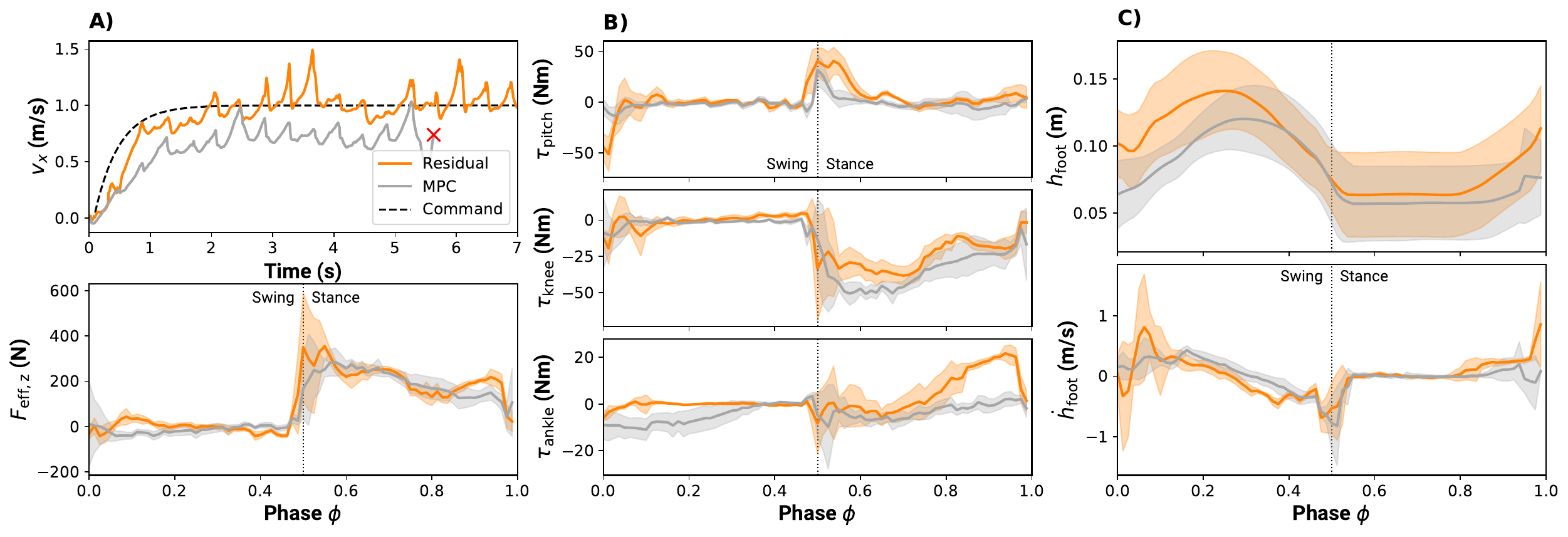}
  \caption{
    Analysis of residual and MPC data from locomotion over uneven terrain at 1.0 m/s.
    \textbf{A)} Velocity tracking and effective vertical commanded force for the right foot.
    \textbf{B)} Sagittal joint torques through the phase. The residual policy exhibits emergent toe-off and heel touchdown behavior from training.
    \textbf{C)} Height and velocity profiles of the center of the foot link during swing and stance.
  }
  \label{fig:residual_forces}
\end{figure*}

\subsection{Performance}
\label{sec:residual_performance}
We demonstrate how the residual architecture outperforms the baseline MPC controller in several metrics.
First, we evaluate the policies with uniformly randomized velocity commands, and plot the 95\% kernel density estimate of the commands with less than 0.25 m/s tracking error, which we consider "achieved".
The resulting boundaries are shown in \autoref{fig:velocity_boundaries}.

The residual policy substantially improves the achievable velocities from the MPC baseline, by roughly 78\% in $v_x$, 12\% in $v_y$, and 9\% in $\omega_z$.
While the performance between the residual and end-to-end policies are similar, as discussed in the previous section, it seems unlikely that the end-to-end policy would be viable for hardware deployment.
The velocities achievable by the end-to-end policy are likely attained by exploiting numeric irregularities in the contact dynamics of the simulator.

Secondly, we explore how the residual architecture adapts the MPC controller for rewards omitted from the MPC cost.
While most of the reward terms in \autoref{tab:rewards} are included in the tracking cost of the MPC controller, a notably difficult constraint to enforce in a model-based setting is self-collision.
Prior works have used simple collision primitives or geometric planning to avoid self-collision, but these methods tend to be heuristic or computationally expensive to solve online \cite{khazoom2024tailoring, ratliff2018riemannian}.
Our MPC formulation contains no self-collision constraints, and we observed that from large commanded $\omega_z$, the knees of the humanoid consistently collided while turning, as shown in \autoref{fig:collision_avoidance}.
The plotted self-collision penalty shows that the MPC controller consistently experiences this self-collision while turning at a constant velocity.

With the residual architecture however, the MPC outputs are modified so that self-collision is avoided entirely.
The residual policy learns to raise the base height of the humanoid torso and adjust its swing trajectory so that the knees never collide in the same manner.
While this is a single illustrative example, the self-collision penalty from the MPC baseline steadily decreases throughout training, approaching zero at convergence.
More broadly, the RL training paradigm is well-suited for costs that are not differentiable or easy to express, while MPC can capture a majority of desirable behavior from a principled, physics-based standpoint.
The residual architecture allows the system to learn behaviors that are difficult to express as smooth constraints in an optimization, and combine the benefits of predictive, model-based control with experience-driven learning.

Next, we investigate the residual policy's adaptability to out-of-distribution scenarios.
Specifically, we consider whether it is able to respond to changes in 1) the requested contact schedule from the MPC and 2) uneven terrain, neither of which was experienced during training.

During inference over flat terrain, we modified the phase parameters of the MPC controller so that a double stance and flight phase would be present in the contact schedule.
Although the MPC prior alone was unable to walk stably with these changes, the residual policy was able to obey these unseen gait commands, as shown in \autoref{fig:contact_scheduling_study}
While prior work has relied on imitating distributions of reference data to adapt gaits, as in \cite{tang2024humanmimic,reske2021imitation}, our residual architecture is able to immediately modify its contact pattern without being trained for it.

We also modify the terrain parameters in IsaacGym, and compare the performance of the baseline MPC policy against the residual architecture.
As shown in \autoref{fig:terrain}, the residual policy is able to navigate unseen terrain with zero fine-tuning or domain randomization.
Despite only being trained on flat, planar ground, the residual policy is capable of navigating uneven terrain with no adaptation, whereas the MPC controller alone immediately fails from incorrect assumptions about the ground height.
For the discrete terrain with more pronounced steps, the feet of the system from the residual policy would occasionally be caught by the environment.
To address this, we modify the swing height parameter in the MPC (see \autoref{fig:mpc_breakdown}) to enforce higher stepping from 0.075 m to 0.15 m.
Although this parameter was also not modified in training, the residual policy immediately adapts to the new swing behavior, and is able to successfully traverse terrain that requires higher clearance.
To our surprise, the residual policy seems surprisingly robust to modifications in the parameters of the underlying MPC controller, allowing for tuning of behavior post-training.

Finally, we demonstrate deploying the proposed residual architecture on hardware on \mithumanoid, as shown in \autoref{fig:hardware}.
The MPC is evaluated at 100 Hz onboard the robot with OSQP~\cite{stellato2020osqp}, and the residual network outputs are directly added to the MPC torques as described in \autoref{sec:blending}.
Video results are available in the attached material.
Out of concern for the integrity of the hardware, experiments were limited in scope to avoid catastrophic failures for the custom system (e.g., testing self-collisions at high angular velocities).
Before deployment, we validated the residual policies in our custom simulator with more accurate contact dynamics.
We observed significant discrepancies in the simulated rollouts, even for the MPC controller alone, likely explaining the sim-to-real gap for hardware deployment.


\subsection{Residual Policy Analysis}
\label{sec:residual_analysis}

In this section, we aim to analyze \textit{when} and \textit{how} the residual network is used during locomotion, especially in relation to the output of the MPC.

We begin by studying under what conditions the residual policy is more "active".
To compare the residual and MPC torques, we use the ratio of their respective magnitudes
\begin{equation}
    \frac{\norm{\boldsymbol{\tau}_{\mathrm{residual}}}}{\norm{\boldsymbol{\tau}_{\mathrm{MPC}}}}.
\end{equation}
While we could also normalize the residual torque against the net joint torque, the MPC and residual torques can oppose each other, masking their true relative contribution.

From \autoref{fig:residual_vs_command}, we see that the contribution from the residual torques is asymmetric with respect to the commands.
Interestingly, the residual torque ratio is positively correlated with the commanded forward velocity.
We hypothesize that the MPC is actually more stable walking backwards than forwards, as was the case with~\cite{khazoom2024tailoring}, and therefore relies less on the residual policy.
When comparing the effect of the commanded lateral and angular velocity, the ratio is high in the second and fourth quadrants (negative $v_y$, positive $\omega_z$ and positive $v_y$, negative $\omega_z$ respectively), and low in the first and third quadrants (positive $v_y$, positive $\omega_z$ and negative $v_y$, negative $\omega_z$ respectively).
We believe this is from the lack of self-collision constraints in the MPC formulation.
With positive/negative lateral velocity and negative/positive angular velocity, the humanoid turns inwards as it moves laterally, risking collision with the forward-protruding knees of the body.
The residual policy needs to actively raise the height of the body and modify the MPC swing trajectory to avoid this collision.
With the other combinations, the humanoid is turning outwards, spreading the knees and stance legs apart, which has a much lower risk of self-collision.

Additionally, we plot the ratio of residual torques as a function of the contact phase in \autoref{fig:residual_analysis}A.
During training, we set the contact parameters such that at $\phi = 0.5$, the commanded contact switches from swing to stance for the right foot, and stance to swing for the left.
Interestingly, we see an increase in the residual torque ratio when the phase is near the switch, right before the feet are expected to make contact with the ground.
Given the robustness of the RL locomotion policies to uneven terrain and imprecise contact timing, we attempt to interpret and characterize the actions of the residual network more closely.

To do so, we compute the normalized dot product between the MPC torques and residual torques
\begin{equation}
    \cos(\theta) = \frac{{\boldsymbol{\tau}_{\mathrm{residual}} \cdot \boldsymbol{\tau}_{\mathrm{MPC}}}} {\norm{\boldsymbol{\tau}_{\mathrm{residual}}} \norm{\boldsymbol{\tau}_{\mathrm{MPC}}}},
\end{equation}
across the phase, as shown in \autoref{fig:residual_analysis}B.
This cosine similarity characterizes the "antagonism" of the residual torque: a value of -1 implies that the residual network exactly opposes the MPC torques in direction, while a value of 1 implies that the residual network is purely additive, scaling the MPC torque by some factor.
When we visualize this as a function of the phase, the residual torques almost \textit{completely oppose} the MPC output near contact switches.
Indeed, the residual torques seem to generally be misaligned with the MPC torques, implying the residual significantly modulates the controller throughout swing and stance.

Finally, we analyze the differences between the residual and baseline MPC policy in \autoref{fig:residual_forces}.
Both controllers are initialized from the same conditions, and commanded to walk forward at 1.0 m/s across uneven terrain.
As discussed previously, the residual policy is capable of successfully navigating uneven terrain, even when it was never encountered during training.
For the MPC controller, however, the accumulated errors from uneven terrain destabilize its solution quality leading to failure shortly after initialization.
For the respective policies, we plot the vertical tracking performance over this terrain, effective vertical ground reaction force (see Appendix \ref{appendix:mapping}), sagittal joint torques, and the foot states.
We notice that the residual policy exhibits significant heel-toe transitions during locomotion, despite not being given rewards to encourage this behavior.
This is reflected in the increased ankle torques at the end of the stance phase and the early takeoff of the foot.
While the vertical force profiles of the two controllers are similar, the residual policy commands a larger reaction force at touchdown, and generally uses more noticeable ankle strategies.
The MPC controller is constrained to touchdown with both the heel and toe at the same time, but the residual policy learns to adapt this, evidenced by the early takeoff of the foot center and ankle torques near the end of stance.
The touchdown velocity of the residual policy also returns to zero in two discernible stages, roughly when the phase is 0.48 and 0.52.
This likely corresponds to the heel and toe touching down separately.
In comparison, the average MPC touchdown velocity is a single sharp peak, and the adherence to a strict contact schedule makes it difficult to adapt to unexpected contacts.

\section{Conclusion}

In this work, we present a unified control architecture that integrates model predictive control with reinforcement learning to achieve robust and adaptable behavior.
We introduced an efficient GPU-parallelized MPC formulation that enables concurrent, in-the-loop policy training at high frequencies, making it feasible to train with optimal controllers in the RL loop on a single desktop computer.
Our analysis shows how the MPC prior can shape and guide the learning process, and we closely study how the residual network can modify MPC outputs to improve robustness under environmental uncertainty.
Finally, we validate the proposed locomotion controller on hardware for \mithumanoid, demonstrating that the combined approach translates from simulation to physical platforms.

While these results are promising, our investigation is only the first step into exploring how predictive controllers and learned policies can be blended together within the RL training process.
Our future work will build upon this control architecture in several ways.
While we strictly limited our scope to only modify the outputs of the predictive controller, we plan to extend the network to also output MPC parameters, such as cost function weights or contact schedules directly.
We also plan to study how the value returned by the MPC solution can be used to estimate uncertainty of the model-based controller, and appropriately balance between the network and MPC.
Large perturbations or errors in model parameters cause spikes in the predicted value, which could serve as a salient signal for the condition of the MPC solution.

Another direction of interest is to design network architectures that mimic the operations required to solve the MPC problem - namely, a factorization step and iterations - to greatly reduce the computational time required.
By carefully constructing a network with permutation invariance and recurrence akin to the ADMM iterations, we anticipate that we can approximate the MPC problem more effectively than with a standard MLP architecture.
With the MPC formulation parallelized on the GPU, data for the individual linear solves, iterations, and full solution would be cheap to collect.


\section{Acknowledgments}
\noindent
The authors would like to thank David Nguyen, Kendrick Cancio, Annika Marschner, and AZ Krebs at the Biomimetic Robotics Lab for their insightful feedback on the saturation and value of \autoref{fig:terrain}.

{\printbibliography}

{\appendix[Joint-to-Cartesian Actuation Mapping]
\label{appendix:mapping}

Having compared and analyzed both residual and MPC outputs in joint space in Section~\ref{sec:residual_analysis}, it is natural to examine how these actions contribute in a more intuitive 3D task space, namely the contact frame located at the foot. Therefore, to better understand how the residual and MPC outputs influence locomotion in task-space coordinates, we transform the output torques from joint space into task space. The details of this transformation are provided in this section.

To analyze the six-dimensional ground reaction wrench (GRW) exerted on each foot, we apply a quasi-static mapping that converts joint torques into task-space GRWs. Specifically, we use the quasi-static equations of motion (i.e., neglecting inertial and Coriolis forces), a common approach used to control task-space ground reaction force~\cite{hong2020real,ding2021representation,di2018dynamic}. This method enables the computation of equivalent joint torques that produce the same effect as if external forces were applied at specified contact locations.
\begin{equation} \label{eq:mapping_1}
    \boldsymbol{\tau} = (\mathbf{J})^\intercal \cdot \mathbf{W},
\end{equation}
where $\boldsymbol{\tau} \in \mathbb{R}^{6}$ denotes the joint torques for each leg, $\mathbf{J} \in \mathbb{R}^{6\times6}$ is the Jacobian matrix mapping joint-space velocities to task-space velocities, and $\mathbf{W}=[f_x,f_y,f_z,m_x,m_y,m_z] \in \mathbb{R}^{6}$ represents the task-space GRW acting at the foot. Given the $\boldsymbol{\tau}$, if the Jacobian matrix is full rank, it can be inverted to uniquely determine $\mathbf{F}$ from $\boldsymbol{\tau}$. For our system, \mithumanoid, each leg has five actuators without having an ankle-roll actuator (i.e., $\boldsymbol{\tau}_{a}=0$), we constrain the direction in which GRWs can be applied:
\begin{equation} \label{eq:mapping_2}
    0=\boldsymbol{\tau}_{a} = \mathbf{a} \cdot \mathbf{W},
\end{equation}
where $\mathbf{a}$ is the corresponding row of $(\mathbf{J})^\intercal$. As a result, during locomotion the sole of the foot typically makes line contact with the ground. Therefore, we specify three translational ground reaction forces $(f_x,f_y,f_z)$ and two ground reaction moments (e.g., $m_y,m_z$) that are orthogonal to the contact line direction, and solve for the ground reaction moment $m_x$: 
\begin{equation} \label{eq:mapping_3}
    m_x = g(f_x,f_y,f_z,m_y,m_z)
\end{equation}
Substituting Eq.~\ref{eq:mapping_3} into Eq.~\ref{eq:mapping_2}, we obtain
\begin{equation} \label{eq:mapping_4}
    \boldsymbol{\tau}^\prime = (\mathbf{J^\prime})^\intercal \cdot \mathbf{W^\prime},
\end{equation}
where $\boldsymbol{\tau}^\prime \in \mathbb{R}^{5}$ denotes the joint torques excluding ankle-roll torque, $\mathbf{W}^\prime = [f_x,f_y,f_z,m_y,m_z] \in \mathbb{R}^{5}$ represents the GRW excluding the line-contact direction, and $\mathbf{J}^\prime = \in \mathbb{R}^{5\times5}$ is the corresponding Jacobian matrix. We now have a square, full-rank matrix $\mathbf{J}^\prime$ that can be inverted to uniquely determine $\mathbf{W}^\prime$ given the joint torques $\boldsymbol{\tau}^\prime$. Substituting $\mathbf{W}^\prime$ into Eq.~\ref{eq:mapping_3}, we can then compute the ankle-roll torque and recover the full six-dimensional wrench $\mathbf{W}$.
}

\end{document}